\documentclass[review]{elsarticle}
\usepackage[table]{xcolor}

\usepackage{hyperref}

\biboptions{authoryear}

\usepackage{booktabs}
\usepackage{graphicx}
\usepackage{caption}
\usepackage{subcaption}
\usepackage{xcolor}
\usepackage{multirow}
\usepackage{tikz}
\usepackage{url}
\usepackage{hyperref}

\makeatletter
\newcommand{\firstlower}{\MakeLowercase}
\newcommand*{\getnamereftext}[1]{%
    \@ifundefined{r@#1}{}{%
    \unexpanded\expandafter\expandafter\expandafter{%
      \expandafter\expandafter\expandafter\@thirdoffive\csname r@#1\endcsname
    }%
  }%
}
\makeatother

\def\checkmark{\tikz\fill[scale=0.4](0,.35) -- (.25,0) -- (1,.7) -- (.25,.15) -- cycle;} 
\begin{document}

\begin{frontmatter}

\title{OdontoAI: A human-in-the-loop labeled data set and an online platform to boost research on dental panoramic radiographs}

\author[ivision]{Bernardo Silva}
\author[ivision]{Laís Pinheiro}

\author[faculdadeodonto]{Brenda Sobrinho}
\author[faculdadeodonto]{Fernanda Lima}
\author[faculdadeodonto]{Bruna Sobrinho}

\author[ifbabarreiras]{Kalyf Abdalla}

\author[faculdadeodontouesb]{Matheus Pithon}

\author[faculdadeodonto]{Patrícia Cury}

\author[ivision]{Luciano Oliveira\texorpdfstring{\corref{correspondingauthor}}{}}
\cortext[correspondingauthor]{Corresponding author}
\ead{lrebouca@ufba.br}

\address[ivision]{Intelligent Vision Research Lab, Federal University of Bahia, Salvador, BA, Brazil}
\address[faculdadeodonto]{Dental School, Federal University of Bahia, Salvador, BA, Brazil}
\address[ifbabarreiras]{Federal Institute of Bahia, Barreiras, BA, Brazil}
\address[faculdadeodontouesb]{Dental School, Southwest State University of Bahia, Jequié, BA, Brazil}

\begin{abstract}

Deep learning has remarkably advanced in the last few years, supported by large labeled data sets.
These data sets are precious yet scarce because of the time-consuming labeling procedures, discouraging researchers from producing them.
This scarcity is especially true in dentistry, where deep learning applications are still in an embryonic stage.
Motivated by this background, we address in this study the construction of a public data set of dental panoramic radiographs.
Our objects of interest are the teeth, which are segmented and numbered, as they are the primary targets for dentists when screening a panoramic radiograph.
We benefited from the human-in-the-loop (HITL) concept to expedite the labeling procedure, using predictions from deep neural networks as provisional labels, later verified by human annotators.
All the gathering and labeling procedures of this novel data set is thoroughly analyzed.
The results were consistent and behaved as expected: At each HITL iteration, the model predictions improved.
Our results demonstrated a 51\% labeling time reduction using HITL, saving us more than 390 continuous working hours.
In a novel online platform, called OdontoAI, created to work as task central for this novel data set, we released 4,000 images, from which 2,000 have their labels publicly available for model fitting.
The labels of the other 2,000 images are private and used for model evaluation considering instance and semantic segmentation and numbering.
To the best of our knowledge, this is the largest-scale publicly available data set for panoramic radiographs, and the OdontoAI is the first platform of its kind in dentistry.

\end{abstract}


\end{frontmatter}

\section{Introduction}

In deep learning-based systems, the number of adjustable parameters of a neural network easily surpasses the million mark, demanding large amounts of data for training.
Most domains, including computer vision, fundamentally rely on supervised learning techniques, which require labeled data to fit the deep learning model weights (\citeauthor{lecun2015deep}, \citeyear{lecun2015deep}).
The labeling procedure depends on human specialists who manually annotate the data according to the application purposes.
This step is crucial and can take up more than 80\% of a machine learning project's time (\citeauthor{wu2021survey}, \citeyear{wu2021survey}).
Consequently, labeled publicly available data sets are valuable resources, and for academic research, they offer the additional benefit of creating benchmarks for model performance comparisons (\citeauthor{Menze2015CVPR}, \citeyear{Menze2015CVPR}; \citeauthor{cordts2016cityscapes}, \citeyear{cordts2016cityscapes}; \citeauthor{wang2018glue}, \citeyear{wang2018glue}).

Many image data sets have promoted progress in the computer vision field (\citeauthor{lecun1998gradient}, \citeyear{lecun1998gradient}; \citeauthor{krizhevsky2009learning}, \citeyear{krizhevsky2009learning}; \citeauthor{deng2009imagenet}, \citeyear{deng2009imagenet}; \citeauthor{lin2014microsoft}, \citeyear{lin2014microsoft}).
The ImageNet (\citeauthor{deng2009imagenet}, \citeyear{deng2009imagenet}) and the COCO (\citeauthor{lin2014microsoft}, \citeyear{lin2014microsoft}) data sets are examples of large-scale and modern labeled image banks.
The construction of these data sets was possible due to crowdsourcing platforms, such as the Amazon Mechanical Turk marketplace (\citeauthor{mechanicalturk}, \citeyear{mechanicalturk}), where one can pay individuals to annotate the images.
Unfortunately, this approach does not work in the medical area, as the field needs qualified professionals to precisely label the data.

In the medical imaging field, some works made large data sets publicly available (\citeauthor{wang2017chestx}, \citeyear{wang2017chestx}; \citeauthor{irvin2019chexpert}, \citeyear{irvin2019chexpert}).
These works relied on medical records or textual reports.
However, when the applications require specific labels, such as bounding boxes or segmentation masks, the data set sizes reduce drastically.
These label annotation procedures are astonishingly onerous, a problem that we can attenuate by using some proposed tools and methods in the literature (\citeauthor{acuna2018efficient}, \citeyear{acuna2018efficient}; \citeauthor{ling2019fast}, \citeyear{ling2019fast}; \citeauthor{liao2021towards}, \citeyear{liao2021towards}), such as the human-in-the-loop (\citeauthor{wu2021survey}, \citeyear{wu2021survey}).

\begin{figure}
    \centering
    \includegraphics[width=\textwidth]{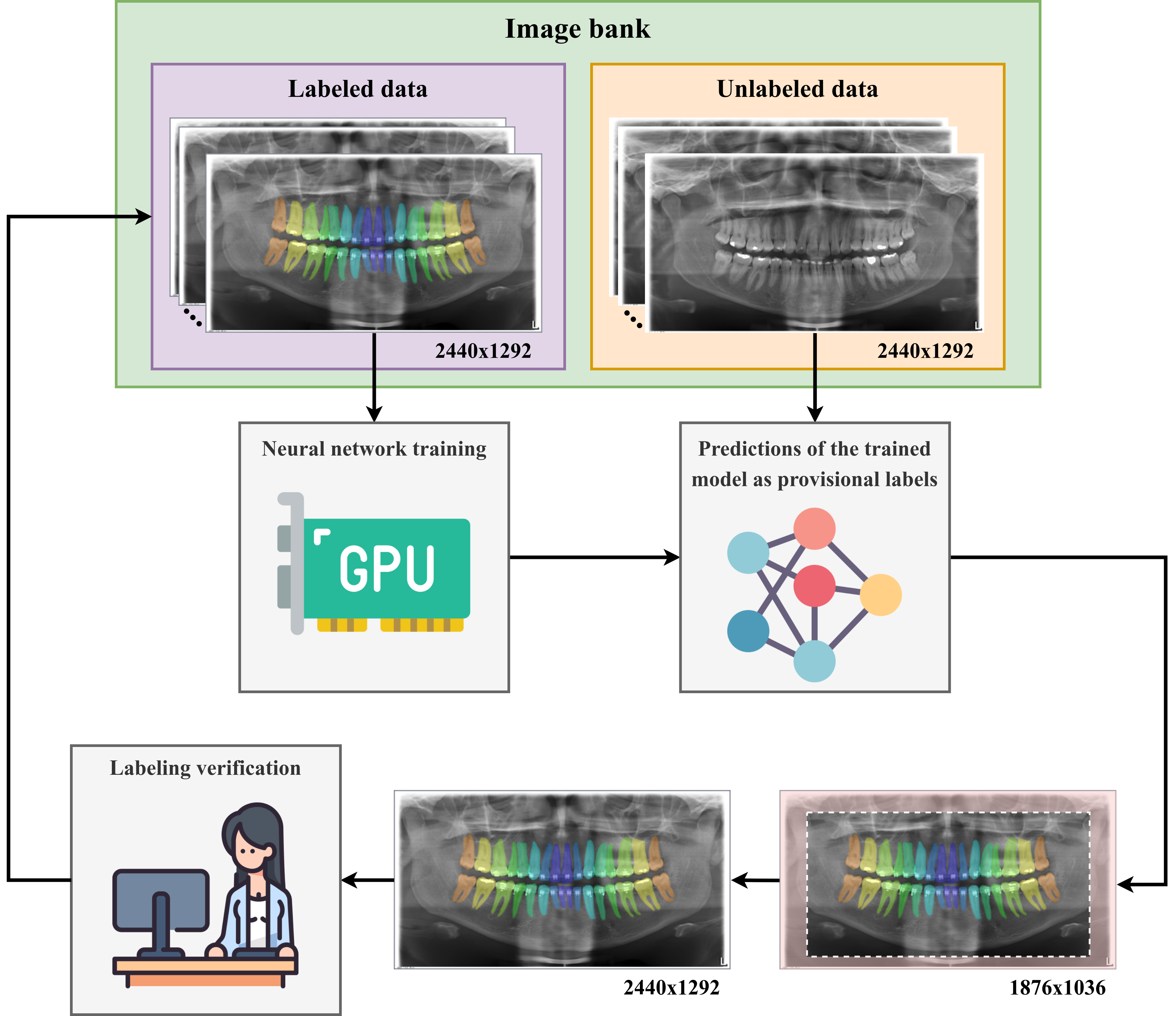}
    \caption{A general HITL diagram for the case of tooth instance segmentation on panoramic radiographs.
    The process works as follows: A neural network is fitted with available labeled data.
    This neural network produces annotations for the unlabeled data, which are later verified by humans.
    The human supervision qualifies the newly labeled data for the model training in subsequent HITL iterations.}
    \label{fig:hitl-concept}
\end{figure}

The human-in-the-loop (HITL) concept is an alternative to crowdsourcing when the latter can not be employed.
HITL pursues to efficiently label the data by combining machine learning models and human supervision, expecting an overall reduction of time and costs (\citeauthor{wu2021survey}, \citeyear{wu2021survey}).
Figure \ref{fig:hitl-concept} displays a generic HITL pipeline with interventional training, using dental panoramic radiographs as example.
In that setup, an initial set of labeled data is used to fit a machine learning model, which later produces annotations for new unlabeled data.
Then, human experts verify (confirm or correct) these annotations, which qualifies them as suitable training and validation data for the next HITL iteration.
After each training iteration, the model performance is expected to improve, increasing the label quality and lessening the verification time.

We employed the HITL method to construct a novel panoramic radiograph data set, \textbf{the OdontoAI Open Panoramic Radiographs (O$^2$PR) data set}\footnote{Our data set will be publicly available upon the article acceptance and publication.}.
The data set aims to fill a gap in the dentistry field, where a large-size and consistently labeled panoramic radiograph data set is lacking, while the deep learning applications are still in an incipient stage compared to other healthcare areas (\citeauthor{schwendicke2020artificial}, \citeyear{schwendicke2020artificial}).
The objects of interest are the teeth, which were segmented and classified since they are the dentists' primary targets when examining a dental panoramic radiograph.
We set three benchmarks for this new data set (\textbf{instance segmentation}, \textbf{semantic segmentation}, and \textbf{numbering}) and released a submission platform (\textbf{the OdontoAI platform}) to boost deep learning research on those images while serving as a task central to other researchers in the field.

\subsection{Imaging in dentistry}

Imaging is a fundamental tool for dentists and oral health experts who use photos, magnetic resonance imaging, ultrasonography, and radiographs, among other techniques, to diagnose patients' conditions and diseases as well as to monitor treatment progressions.
As for the radiographs, the most common ones are the periapical, the bitewing, and the panoramic.
In a comprehensive study, \citeauthor{silva2018automatic} (\citeyear{silva2018automatic}) reviewed several papers on segmenting teeth in radiographs. They observed that most works prior to 2018 neglected dental panoramic radiographs, probably due to the challenging nature of these images and the limited performance of the unsupervised segmentation methods.

\begin{figure}[tp]
    \centering
    \includegraphics[width=\textwidth]{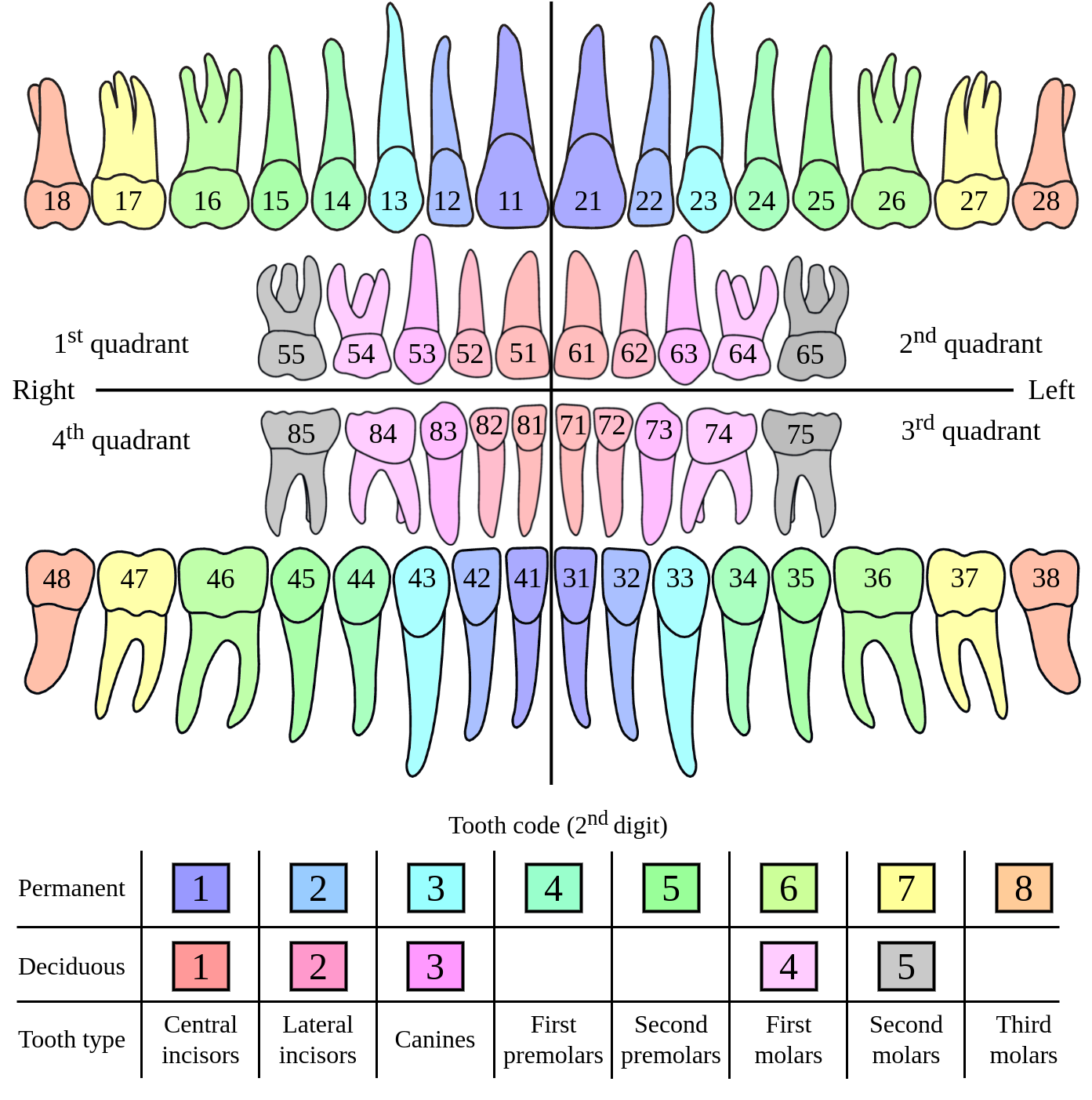}
    \caption{The illustration of FDI World Dental Federation notation.
    The system designates each tooth by a two-digit number, in which the first digit determines the quadrant and the dentition type (permanent or deciduous), and the second digit determines the tooth type.
    We added a custom color code to identify each tooth in our qualitative results.
    Source: Adapted from \citeauthor{pinheiro2021numbering} (\citeyear{pinheiro2021numbering}).}
    \label{fig:fdi-numbering}
\end{figure}

When examining a panoramic radiograph, radiologists usually focus on the teeth, using them as landmarks to analyze the image and report their findings.
The specialists also register the patients' missing teeth and the silhouettes of the existing ones are helpful for forensic identification.
Similar processes occur in computer-aided diagnostic tools.
The dentists use a numeric notation in their written reports, as well as in their daily routines, to avoid citing the full tooth name and expedite communication.
The most common tooth numbering system is the FDI World Dental Federation notation, which represents each tooth by a two-digit number.
The first digit specifies the quadrant and the dentition type (permanent or deciduous), while the second digit specifies the tooth type.
In this work, we employ the FDI notation together with an additional custom color code system used to illustrate the qualitative results.
We illustrate both systems in Figure \ref{fig:fdi-numbering}, and we refer as ``numbering'' the act of identifying each tooth using the FDI notation\footnote{For simplicity's sake, we disregarded the supernumerary teeth in our analyses.}.

\subsection{Literature review}

\citeauthor{silva2018automatic} (\citeyear{silva2018automatic}) were pioneers in applying deep learning to segment teeth on panoramic radiographs.
They employed a Mask R-CNN (\citeauthor{he2017mask}, \citeyear{he2017mask}) trained over binary masks that separated the teeth from the background and showed that their approach outperformed traditional solutions to the task.
The authors also made their data public under the name UFBA-UESC Dental Image Data Set\footnote{\label{foot:links-datasets}The instructions on how to request the mentioned data sets are at:\\
https://github.com/IvisionLab/dental-image (UFBA-UESC Dental Images)\\
https://github.com/IvisionLab/deep-dental-image (UFBA-UESC Dental Images Deep)\\
https://github.com/IvisionLab/dns-panoramic-images (DNS Panoramic Images)\\
https://github.com/IvisionLab/dns-panoramic-images-v2 (DNS Panoramic Images v2)}, which proved to be a valuable resource as it has been extensively used by many works
(\citeauthor{koch2019accurate}, \citeyear{koch2019accurate}; \citeauthor{zhao2020tsasnet}, \citeyear{zhao2020tsasnet}; \citeauthor{oliveira2020truly}; \citeyear{oliveira2020truly}; \citeauthor{chen2021mslpnet}, \citeyear{chen2021mslpnet};  \citeauthor{cui2021toothpix}, \citeyear{cui2021toothpix}; \citeauthor{hsu2021deepopg}, \citeyear{hsu2021deepopg}).
In an extension of \citeauthor{silva2018automatic}'s work, \citeauthor{jader2018deep} (\citeyear{jader2018deep}) segmented tooth instances on radiographs of the UFBA-UESC Dental Image Data Set also using Mask R-CNN, though not numbering them.
In order to perform instance segmentation, the authors manually modified 276 binary masks from the original data set that separated the teeth from the background.
This modification produced labels that disregarded tooth overlapping, but their results surpassed the preceding ones pronouncedly.
The authors made their data publicly available under the name UFBA-UESC Dental Image \textbf{Deep} Data Set\textsuperscript{\ref{foot:links-datasets}}.

\citeauthor{silva2020study} (\citeyear{silva2020study}) advanced the field by segmenting and numbering tooth instances.
They conducted a benchmark with 543 radiographs from the UFBA-UESC Dental Image Data Set, modifying the original masks the same way \citeauthor{jader2018deep} (\citeyear{jader2018deep}) did, also incorporating numbering labels to the permanent teeth.
The benchmark assessed the performance of four end-to-end instance segmentation neural network architectures that achieved state-of-the-art performance on the COCO data set: Mask R-CNN, PANet (\citeauthor{liu2018path}, \citeyear{liu2018path}), Hybrid Task Cascade (HTC) (\citeauthor{chen2019hybrid}, \citeyear{chen2019hybrid}), and Cascade Mask R-CNN backboned by a ResNeSt (\citeauthor{zhang2020resnest}, \citeyear{zhang2020resnest}).
The benchmark winner architecture was the PANet, but the authors concluded that all architectures had satisfactory performances on the task.
Their data are publicly available under the name DNS Panoramic Images\textsuperscript{\ref{foot:links-datasets}}.

Lastly, \citeauthor{pinheiro2021numbering} (\citeyear{pinheiro2021numbering}) labeled from scratch a subset of 450 radiographs from the UFBA-UESC Dental Image Data Set (\citeauthor{silva2020study}, \citeyear{silva2020study}) considering tooth overlapping and deciduous teeth, topics neglected by previous studies.
The authors refined the Mask R-CNN prediction through the aid of the PointRend module (\citeauthor{kirillov2020pointrend}, \citeyear{kirillov2020pointrend}).
They demonstrated that it is feasible to accurately number and segment permanent and deciduous teeth through end-to-end deep learning solutions and that the PointRend module was more beneficial for segmenting more complex-shaped teeth.
They named their data set DNS Panoramic Image \textbf{v2} and made it publicly available\textsuperscript{\ref{foot:links-datasets}}.

\citeauthor{tuzoff2019tooth} (\citeyear{tuzoff2019tooth}) proposed a two-stage solution for detecting and numbering teeth.
In the first stage, a Faster R-CNN network (\citeauthor{ren2015faster}, \citeyear{ren2015faster}) detects the teeth without numbering them. The detections define the areas used to generate the crops for the next stage.
These crops are bigger than the tooth bounding boxes, which adds location context, easing the classification task.
In the second stage, a VGG-16 classification network (\citeauthor{simonyan2014very}, \citeyear{simonyan2014very}) takes these crops as inputs and classifies the teeth.
In total, the experiments relied on 1572 not publicly available images, labeled with bounding boxes by specialists.
\citeauthor{leite2021artificial} (\citeyear{leite2021artificial}) proposed a two-stage solution to perform segmentation and numbering.
In the first stage, a DeepLabv3 network (\citeauthor{chen2017rethinking}, \citeyear{chen2017rethinking}), backboned by a ResNet-101 (\citeauthor{he2016deep}, \citeyear{he2016deep}), segments 16 tooth classes (two incisors, one canine, two premolars, three molars for each dental arch).
In the second stage, a fully convolutional network (FCN) (\citeauthor{long2015fully}, \citeyear{long2015fully}) refines the segmentation predictions.
For their experiments, the authors employed 153 panoramic radiographs labeled by an expert from a private data set.
The two prior solutions had the inherent drawback of not allowing end-to-end training.

\begin{table}[t]
\centering
\caption{Main features of the previous works' data sets and ours. To the best of our knowledge, this work's data set is the largest on tooth instance segmentation and numbering of dental panoramic radiographs.}
\label{tab:datasets}
\resizebox{\textwidth}{!}{%
\begin{tabular}{@{}cccccccc@{}}
\toprule
Authors & \# Radiographs & Detection  & Numbering  & Segmentation & Image dimensions & Availability & Annotators \\ \midrule
\cite{silva2018automatic}    & 1,500 &            &            & \checkmark & $1,991\times1,127$ & Public  & Lay people \\
\cite{jader2018deep}         & 276  & \checkmark &            & \checkmark & $1,991\times1,127$ & Public  & Lay people \\
\cite{tuzoff2019tooth}       & 1,572 & \checkmark & \checkmark &            & N.A.             & Private & Experts    \\
\cite{silva2020study}        & 543  & \checkmark & \checkmark & \checkmark & $1,991\times1,127$ & Public  & Students   \\
\cite{chung2021individual}   & 818  & \checkmark & \checkmark &            & Several          & Private & Experts    \\
\cite{leite2021artificial}   & 153  & \checkmark & \checkmark & \checkmark & $2,880\times1,504$ & Private & Expert     \\
\cite{pinheiro2021numbering} & 450  & \checkmark & \checkmark & \checkmark & $1,876\times1,036$ & Public  & Mixed      \\
\cite{krois2021impact}       & 5,008 & \checkmark & \checkmark &            & N.A.             & Private & Mixed      \\
\cite{panetta2021tufts}      & 1,000 & \checkmark & \checkmark & \checkmark & $1,615\times840$         & Public  & Mixed      \\ \midrule
\textbf{We}                & 4,000 & \checkmark & \checkmark & \checkmark & $2,440\times1,292$ & Public  & Mixed      \\ \bottomrule
\end{tabular}%
}
\end{table}

\citeauthor{chung2021individual} (\citeyear{chung2021individual}) developed a new method for detecting and classifying teeth on panoramic radiographs. 
Firstly, through linear regression, the method localizes 32 points, each representing a single permanent tooth in an adult mouth regardless of its presence, automatically numbering them.
In the second and final stage, the point coordinates are refined, and the tooth bounding boxes are predicted in a cascade manner.
This approach ignores deciduous and supernumerary teeth.

\citeauthor{krois2021impact} (\citeyear{krois2021impact}) examined the impact of image context on tooth classification.
The authors showed that a model performance can significantly increase with additional context around the tooth bounding boxes.
They confirmed this fact by training and evaluating ResNet-34 networks to classify teeth with different contexts on a private data set comprising 5004 dental panoramic radiographs in total.
More than 50 annotators were involved in labeling this large amount of data.
Finally, \citeauthor{panetta2021tufts} (\citeyear{panetta2021tufts}), constructed and published a multimodal dental panoramic radiograph data set.
The data set comprises 1,000 radiographs and labels for tooth instance segmentation, abnormalities, eye-tracking, and textual description.
The authors established some baselines only for semantic segmentation.

There are several gaps and shortages in the literature current state.
The main issues come from the data, as many researchers collect and label radiographs only for their studies, with custom labeling standards.
Consequently, the researchers' precious time is wasted at each new study, and various metrics are employed, hindering any possibility of comparing the proposed solutions' performance.
In our work, we tackled those problems by (i) introducing a large-scale, fine-labeled, and high-variability data set for tooth segmentation and numbering, comprising 4,000 dental panoramic radiographs built upon the HITL concept and (ii) releasing an online platform for benchmarking solutions to work as task central for instance segmentation, semantic segmentation, and numbering.
Table \ref{tab:datasets} displays the main features of the reviewed work here in comparison with ours.

\subsection{Paper outline}

This paper's structure follows, in chronological order, the several steps required to construct our data set using the HITL concept and analyze their outcomes.
Table \ref{tab:section-goals} summarizes all these steps, mentioning the number of used images at each step as well as the tasks benchmarked on our online platform.
We detail and discuss all those topics in the following sections.

First, we established the standards for \textbf{\firstlower{\getnamereftext{section:manual-labeling}}} (Section \ref{section:manual-labeling}) used to annotate the data for the first HITL iteration while constructing the test data set.
We then stipulated the parameters and conventions for the \textbf{\getnamereftext{section:hitl-setup}} (Section \ref{section:hitl-setup}).
A benchmark for
\textbf{selecting the deep learning architecture for the HITL scheme}
(Section \ref{section:benchmark}) aimed to choose the most suitable instance segmentation architecture for our experiments.
We proceed with the \textbf{\getnamereftext{section:hitl-labeling}} (Section \ref{section:hitl-labeling}) protocol followed by our annotators.
We conducted several analyses to evaluate the HITL benefits, starting with \textbf{\firstlower{\getnamereftext{section:results-on-validation-sets}}} (Section \ref{section:results-on-validation-sets}), 
\textbf{model results on HITL data} (Section \ref{section:results-on-hitl-data}), and
\textbf{\firstlower{\getnamereftext{section:results-on-test-data}}} (Section \ref{section:results-on-test-data}).
The \textbf{\firstlower{\getnamereftext{section:numbering-analysis}}} (Section \ref{section:numbering-analysis}) assessed the network performance evolution on a less painstaking task.
We also perform a \textbf{\firstlower{\getnamereftext{section:labeling-time-analysis}}} (Section \ref{section:labeling-time-analysis}), comparing the manual and HITL labeling.
We investigate the main \textbf{\getnamereftext{section:bottlenecks}} (Section \ref{section:bottlenecks}) for faster labeling verification.
The \textbf{\firstlower{\getnamereftext{section:qualitative-analysis}}} (Section \ref{section:qualitative-analysis}) follows the quantitative analyses, comparing the model predictions with the HITL outcomes.
Finally, we detail the \textbf{\firstlower{\getnamereftext{sec:inst-segm-task}}} (Section \ref{sec:inst-segm-task}), \textbf{\firstlower{\getnamereftext{sec:seman-segm-task}}} (Section \ref{sec:seman-segm-task}), and \textbf{\firstlower{\getnamereftext{sec:numbering-task}}} (Section \ref{sec:numbering-task}) along with the established evaluation protocols and baselines used in the OdontoAI platform.

\begin{table}[t]
\centering
\caption{The steps, along with the number of used images, required to construct our data set and to analyze the HITL outcomes. This paper is structured to detail each step in its corresponding section.}
\label{tab:section-goals}
\resizebox{\textwidth}{!}{%
\begin{tabular}{@{}cp{6cm}cp{10cm}@{}}
\toprule
Section & Step & \# Images & Goal description \\ \midrule
\ref{section:manual-labeling} & \getnamereftext{section:manual-labeling} & 850  & Establish the manual labeling standards used in the first HITL iteration and to construct the test data set. \\
\ref{section:hitl-setup} & \getnamereftext{section:hitl-setup} & - & Stipulate the parameters and conventions for our HITL scheme. \\
\ref{section:benchmark} & \getnamereftext{section:benchmark} & 450 & Select through a benchmark the best instance segmentation neural network architecture for our HITL scheme. \\
\ref{section:hitl-labeling} & \getnamereftext{section:hitl-labeling} & 3,150 & Label images using the HITL concept to speed up the labeling process. \\
\ref{section:results-on-validation-sets} & \getnamereftext{section:results-on-validation-sets} & 90, 180, 360, 720 & Verify the improvement of network results on validation data \\
\ref{section:results-on-hitl-data} & \getnamereftext{section:results-on-hitl-data} & 450, 900, 1800 & Assess the network results on the large and unbiased data \\
\ref{section:results-on-test-data} & \getnamereftext{section:results-on-test-data} & 400 & Assess and analyze the network results on the test data set. \\
\ref{section:numbering-analysis} & \getnamereftext{section:numbering-analysis} & 400 & Assess the network performances on numbering, a less painstaking task. \\
\ref{section:labeling-time-analysis} & \getnamereftext{section:labeling-time-analysis} & 50 per annotator & Measure the HITL labeling speed up gain against manual labeling. \\
\ref{section:bottlenecks} & \getnamereftext{section:bottlenecks} & 3,150 & Determine the main bottlenecks of the HITL labeling. \\
\ref{section:qualitative-analysis} & \getnamereftext{section:qualitative-analysis} & 400 & Visually inspect the network predictions and HITL outcomes. \\
\ref{sec:inst-segm-task} & \getnamereftext{sec:inst-segm-task} & 4,000 & Establish the instance segmentation evaluation protocol and baselines for the OdontoAI platform. \\
\ref{sec:seman-segm-task} & \getnamereftext{sec:seman-segm-task} & 4,000 & Establish the semantic segmentation evaluation protocol and baselines for the OdontoAI platform. \\
\ref{sec:numbering-task} & \getnamereftext{sec:numbering-task} & 4,000 & Establish the numbering task evaluation protocol and baselines for the OdontoAI platform. \\

\bottomrule
\end{tabular}%
}
\end{table}

\subsection{Contributions}

Our data set reviews, improves, and extends the UFBA-UESC Dental Image Data Set.
We started from a previous study from ours, the DNS Panoramic Images v2 (\citeauthor{pinheiro2021numbering}, \citeyear{pinheiro2021numbering}), in which we manually segmented and numbered the teeth of 450 radiographs from the UFBA-UESC Dental Image Data Set.
In the current work, we benchmarked several instance segmentation neural networks trained from these images to fix the architecture for the HITL scheme, which we adopted to speed up the annotation process.
At each HITL iteration, with the available data, we trained a neural network whose predictions on unlabeled images would later be verified by our annotators.
This process resulted in our new data set, so-called \textbf{O$^2$PR}.
The data set comprises 4,000 images, from which 2,000 have their labels publicly available.
We go a step further and release a platform where researchers can submit their solutions for three different benchmarks that employ the labels of the remaining 2,000 radiographs.
This access restriction is beneficial as it will decrease assessment biases.

\section{Data set construction}

We built the \textbf{O$^2$PR data set} upon 1,493 radiographs from the UFBA-UESC Dental Image Data Set (we discarded seven images from the original 1,500 due to duplicates) and 2,507 additional images, totaling 4,000 radiographs.
All radiographs came from a database of images\footnote{The National Commission for Research Ethics (CONEP) and the Research Ethics Committee (CEP) authorized the use of the radiographs in research under the report number 646.050/2014.} acquired by an ORTHOPHOS XG 5/XG 5 DS/Ceph device.
\citeauthor{silva2018automatic} (\citeyear{silva2018automatic}) grouped the original 1,493 images into ten radiograph categories, according to the presence of dental appliances, restorations, and 32 teeth.
Two supplementary categories are exclusive for radiographs with dental implants and mouths with more than 32 teeth.
This categorization demonstrated the high variability of the data.
Following this categorization, we grouped the remaining radiographs of the image bank and noted that the database category proportions differed from the original 1,493 images subset.
For instance, the radiographs of category 1 were too oversampled in the UFBA-UESC Dental Image Data Set while images of category 8 were subsampled.
We conducted our HITL procedure so that, at the end of it, the 4,000 images' category proportions would be similar to the image database's.
Table \ref{tab:datasets-features-and-proportions} summarizes the \textbf{O$^2$PR data set} according to the number of images per category in the original data set and the newly selected ones for the HITL.

\begin{table}[t]
\centering
\caption{Summary of the UFBA-UESC Dental Image and O$^2$PR data sets according to the number of images per radiograph category. We conducted the HITL procedure so the OdontoAI's radiograph category proportions were similar to the image database's.}
\label{tab:datasets-features-and-proportions}
\resizebox{\textwidth}{!}{%
\begin{tabular}{@{}cccccc@{}}
\toprule
Category & 32 Teeth        & Restorations      & Dental Applicance      & UFBA-UESC Dental Image & \textbf{O$^2$PR} \\ \midrule
1        & \checkmark      & \checkmark        & \checkmark             & 73                & 93           \\
2        & \checkmark      & \checkmark        &                        & 219               & 438          \\
3        & \checkmark      &                   & \checkmark             & 45                & 110          \\
4        & \checkmark      &                   &                        & 138               & 274          \\
5        & \multicolumn{3}{c}{Radiographs contaning dental implant(s)}  & 120               & 228          \\
6        & \multicolumn{3}{c}{Radiographs contaning more than 32 teeth} & 169               & 335          \\
7        &                 & \checkmark        & \checkmark             & 114               & 420          \\
8        &                 & \checkmark        &                        & 455               & 1804         \\
9        &                 &                   & \checkmark             & 45                & 93           \\
10       &                 &                   &                        & 115               & 205          \\ \midrule
\multicolumn{4}{c}{Total}                                               & 1493              & 4000         \\ \bottomrule
\end{tabular}%
}
\end{table}

\subsection{Manual labeling}
\label{section:manual-labeling}
The \textbf{O$^2$PR data set} includes 850 manually annotated images, from which 650 have their labels public, while the labels of other 200 images remain private for model assessments on the OdontoAI platform.
We started from a former work from ours (\citeauthor{pinheiro2021numbering}, \citeyear{pinheiro2021numbering}), in which four students labeled a subset of 450 images from the UFBA-UESC Dental Image Data Set.
In that work, the image selection was random but respected the original data set's category proportions.
The students were two dentistry undergraduates and two STEM graduates experienced in the research of tooth segmentation and numbering on panoramic radiographs.
An experienced radiologist supervised the students' work.
Each student labeled about a fourth of the images using the COCO Annotator software and its polygon tool (\citeauthor{cocoannotator}, \citeyear{cocoannotator}).
The annotators should click on the tooth borders precisely as possible to delineate the teeth' outline, being expected crisper segmentations on sharp and well-focused images.
On blurry images or regions, the students should picture the tooth contours based on their anatomical structure and label them accordingly, except when there was solid evidence for not doing so.
some criteria were defined as to be the standards for the labeling procedure:

\begin{enumerate}[(a)]
    \item Implants should not be labeled;
    \item Prostheses should be incorporated into the tooth instances if they are associated with a single tooth root. If not, only the prothesis portions related to the tooth root in question should be considered;
    \item The palatine root of molars should be segmented, even if the spot is blurry;
    \item Restorations should be fused to the corresponding tooth instances;
    \item Dental appliances should be ignored. For labeling, the annotators should picture the tooth silhouettes when apparatuses, such as brackets and metal rings, blocked the visualization;
\end{enumerate}

\begin{figure}
    \centering
    \begin{subfigure}[b]{0.19\textwidth}
        \centering
        \includegraphics[width=\textwidth]{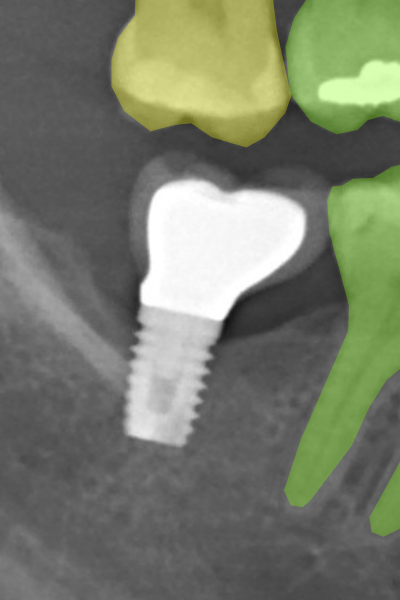}
        \caption{Implants.}
        \label{fig:a-implants}
    \end{subfigure}
    \hfill
    \begin{subfigure}[b]{0.19\textwidth}
        \centering
        \includegraphics[width=\textwidth]{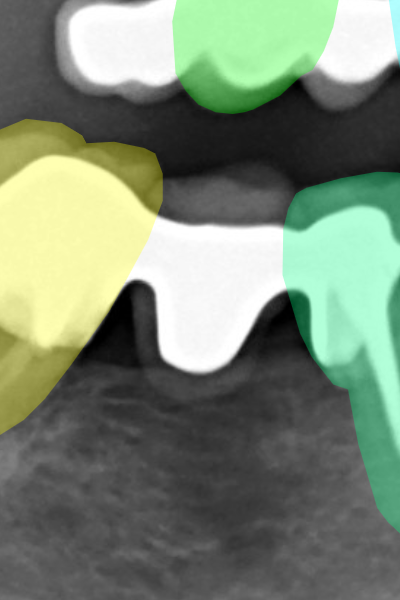}
        \caption{Protheses.}
        \label{fig:b-protheses}
    \end{subfigure}
    \hfill
    \begin{subfigure}[b]{0.19\textwidth}
        \centering
        \includegraphics[width=\textwidth]{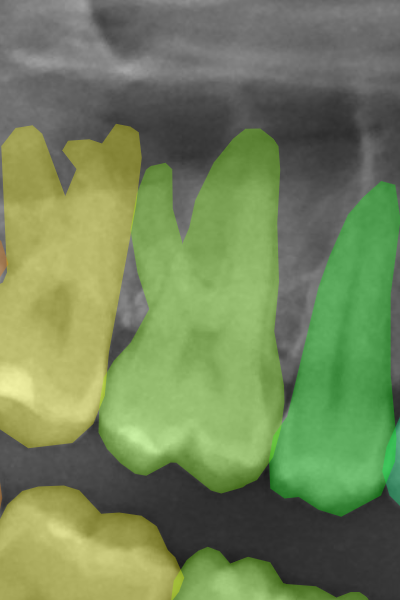}
        \caption{Molar roots.}
        \label{fig:c-molar-roots}
    \end{subfigure}
    \hfill
    \begin{subfigure}[b]{0.19\textwidth}
        \centering
        \includegraphics[width=\textwidth]{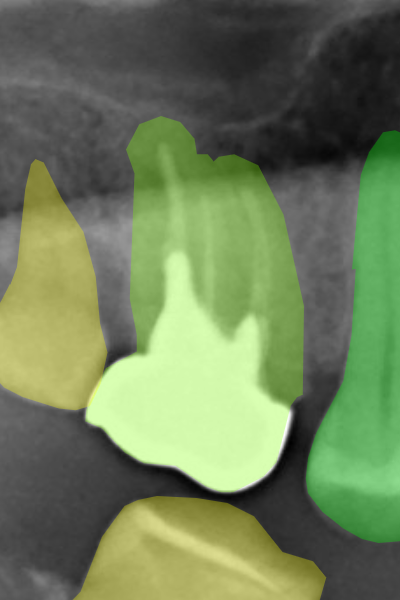}
        \caption{Restorations.}
        \label{fig:d-restorations}
    \end{subfigure}
    \hfill
    \begin{subfigure}[b]{0.19\textwidth}
        \centering
        \includegraphics[width=\textwidth]{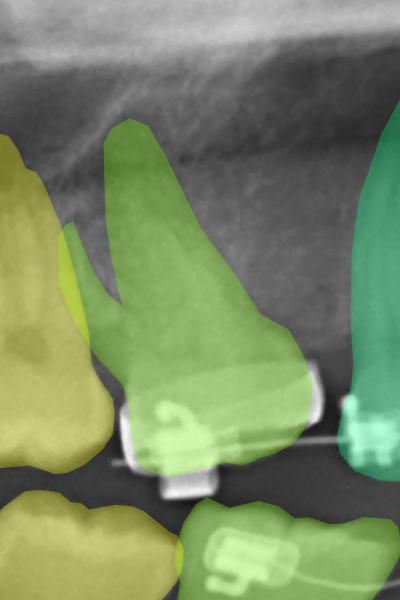}
        \caption{Appliances.}
        \label{fig:e-appliances}
    \end{subfigure}
    \caption{Label samples of the employed criteria for annotating implants, prostheses, molar roots, restorations, and dental appliances. In general, the labels should be more refined on sharp and well-focused images, while in blurry images, the annotators should rely more on the tooth anatomical structures.}
    \label{fig:annotation-criteria}
\end{figure}

Figure \ref{fig:annotation-criteria} displays corresponding label samples of the aforementioned criteria.
We followed the same criteria to label 400 additional images (40 per radiograph category).
These images compounded our test set for assessing the neural networks trained at each HITL iteration.

\subsection{HITL setup}
\label{section:hitl-setup}

\begin{figure}
    \centering
    \includegraphics[height=0.9\textheight]{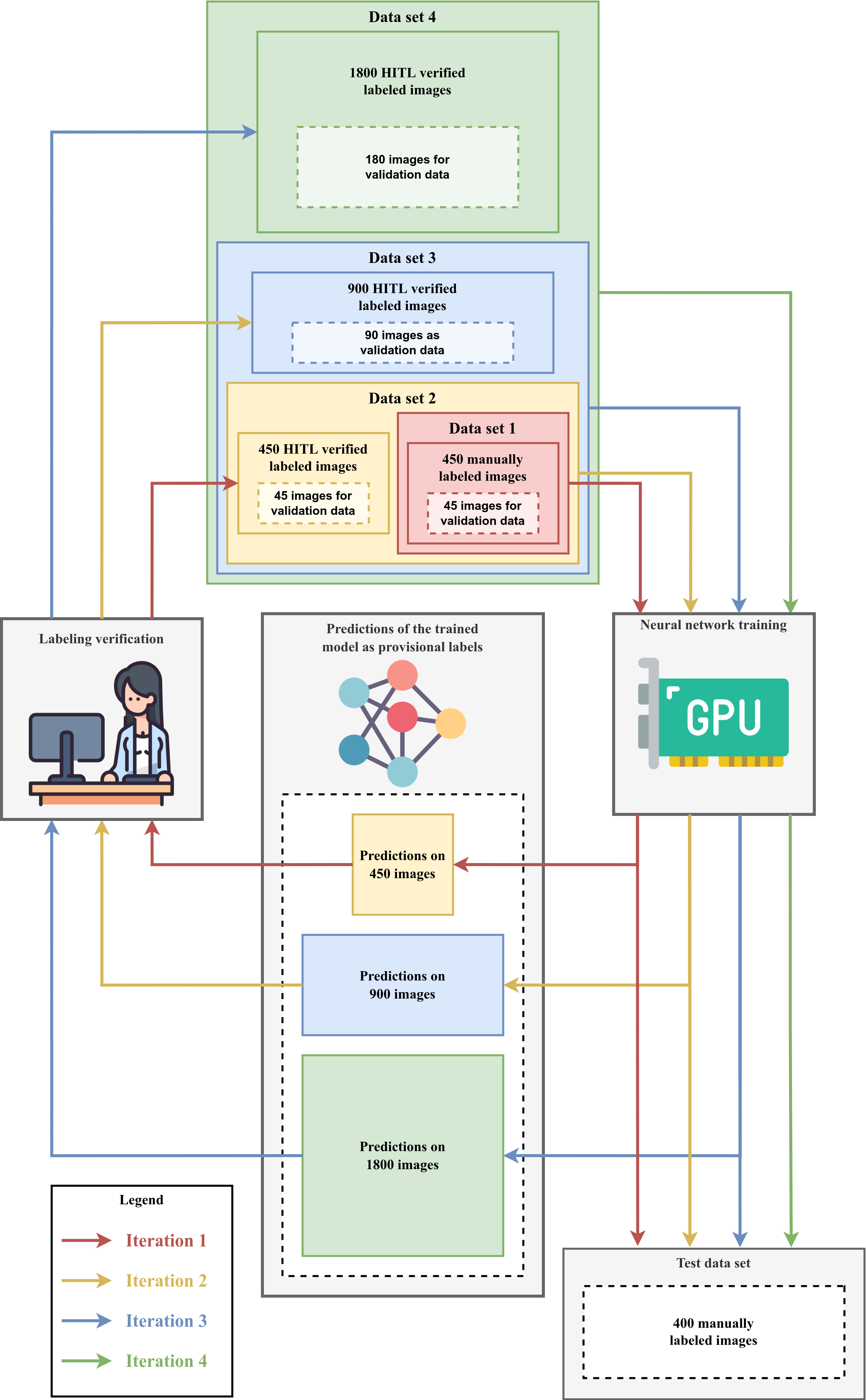}
    \caption{Our human-in-the-loop setup: we trained a neural network with the available labeled radiographs and verify the model predictions on unlabeled images. In this setup, we doubled the number of labeled images at each iteration.}
    \label{fig:hitl-setup}
\end{figure}

Our HITL methodology consisted of the cycle depicted in Figure \ref{fig:hitl-setup}: We trained a network with the available labeled radiographs and, subsequently, used its predictions as provisional labels for a new set of images.
Our annotators verified these labels, which were incorporated in the next iteration into our training and validation sets for a new neural network training, restarting the cycle.

The adopted methodology demanded the setting of some parameters and conventions.
For instance, we had to define the image subset size of the HITL labeled images and how to conduct the neural network training.
A reasonable choice to consider was to label a single image using the model prediction and, subsequently, fine-tune the current weights of the neural network using the available labeled radiographs incremented by the newly available one.
We did not follow this approach due to theoretical concerns and practical issues.
The theoretical concerns were primarily due to the possibility of bias induction towards the first employed images and labels because, at each cycle step, the neural network training would start from a state where the weights would better suit those images.
The practical issues lay in the fact that our annotators should verify the model predictions in the same proportions.
Our annotators had different time-availability and daily routines.
It was impractical for them to verify the HITL annotations in small, consistent cycles (\textit{e.g.}, the first annotator verifies the predictions over one image, the second annotator verifies the predictions over a second image, the third annotator validates another one, and so on).
Letting the annotators verify the labels in any order would introduce untraceable biases, hindering us from performing in-depth analyses of the HITL outcomes.

In order to avoid any of the concerns mentioned above, we chose to verify the instance segmentation predictions for large sets of images at each cycle.
Our setup labeled the same amount of images at each HITL iteration as the number of images used for training and validation.
This way, we began with 450 images split in training (405 images) and validation (45 images), doubling the set sizes at each cycle: The first iteration used 450 images (405 for training and 45 for validation), the second 900 images (810 for training and 90 for validation), continuing until we reached the fourth and final iteration containing 3600 images (3240 for training and 360 for validation).
We hoped with this setup that, at each HITL iteration, we could perceive an improvement in the neural network performance, increasing the labeling quality.

\subsection{Selecting the deep learning architecture for the HITL scheme}
\label{section:benchmark}

We conducted a benchmark with several state-of-the-art instance segmentation neural network architectures to define the model to be used in the HITL scheme.
We selected the available architectures with the highest mean average precision (mAP) values on the COCO 2017 validation set.
The mAP metric is a synonym for mAP@0.5:0.05:0.95, indicating that mAP is the mean value of the ten average precision (AP) scores with true positive thresholds of 0.5 up to 0.95 (inclusive) in steps of 0.05.
The AP means, for each class, the area under the curve (AUC) of the precision-recall graph, where a true positive is computed when the prediction has an intersection over union (IoU) with a ground truth segmentation larger than the considered threshold.
Usual AP metrics include AP50 (AP with 0.5 threshold value) and AP75 (threshold of 0.75).
A threshold value equal to or larger than 0.85 is stringent, as well the final mAP metric.
Throughout this study, we used mAP as the primary metric in most of our experiments and analyses.
In the end, our benchmark comprised seven architectures in total: A conventional Mask R-CNN (\citeauthor{he2017mask}; \citeyear{he2017mask}), backboned by a ResNeXt-101-64x4d; Cascade Mask R-CNN (\citeauthor{cai2019cascade}, \citeyear{cai2019cascade}), backboned by a ResNeXt-101-64x4d; Mask R-CNN backboned by a ResNeSt-101 (\citeauthor{zhang2020resnest}, \citeyear{zhang2020resnest}); Cascade Mask R-CNN with Deformable Convolutional Networks (DCN) (\citeauthor{dai2017deformable}, \citeyear{dai2017deformable}) backboned by a ResNeXt-101-64x4d; Cascade Mask R-CNN  backboned by a ResNeSt-101 (\citeauthor{zhang2020resnest}, \citeyear{zhang2020resnest}); Hybrid Task Cascade (HTC) (\citeauthor{chen2019hybrid}, \citeyear{chen2019hybrid}) with DCN backboned by a ResNeXt-101-64x4d; and DetectoRS (\citeauthor{qiao2021detectors}, \citeyear{qiao2021detectors}) with HTC head backboned by a ResNet-50.

Each selected architecture introduced or adopted appropriate techniques to boost its COCO instance segmentation benchmark metrics.
Mask R-CNN was the first architecture to extend the Faster R-CNN to instance segmentation by adding a mask branch.
It also introduced the RoiAlign, a quantization-free layer, and employed a Feature Pyramid Network (FPN) (\citeauthor{lin2017feature}, \citeyear{lin2017feature}).
The Cascade R-CNN or Cascade Mask R-CNN demonstrated the benefit of using a sequence of detectors with increasing IoU thresholds.
DCNs are neural network modules that enhance the CNNs capabilities on transformation modeling, adding only a small computational overhead.
The ResNeSt backbone stacks modular ResNet-like blocks that can attend to different feature-map groups.
HTC's main contribution is a framework that interweaves the detection and segmentation tasks in a cascade fashion.
Finally, DetectoRS improves object detection with two strategies: (i) Recursive Feature Pyramid, which modifies the FPNs through extra feedback connections to the bottom-up backbone layers, and (ii) Switchable Atrous Convolution, which are convolution operations with different atrous rates whose results are aggregated by switch functions.
It is worth emphasizing that the benchmark ultimate goal was not to fairly compare methods and techniques (as the network and backbone sizes could vary significantly) but rather to specify a solid and reliable architecture to be used in our HITL scheme.

The benchmark protocol and the neural network training procedure for the HITL iterations were the same: It consisted in training each architecture for 150 epochs, with 90\% of the available data as training set and 10\% as validation set.
We cropped all images to the reduced dimensions of 1876 $\times$ 1036 (159 pixels from the top and horizontally centered) to improve the network performances.
These numbers and methodology came by roughly removing 80\% of the extent between the outermost segmentations and the image borders of the 450 firstly labeled radiographs.
This cropping may exclude tooth parts, or even the entire instances, hindering some applications, but, in the HITL, the human supervisor can catch these eventualities and correct them.

Each network performance was measured at the end of each epoch, and we saved the weights corresponding to the highest attained segmentation mAP (early stopping).
The optimizer was the stochastic gradient descent (SGD) with a 0.9 momentum value and no weight decay.
We trained the models with eight Tesla V100 16GB GPUs with a batch size of 8 (one sample per GPU).
We employed a linear warm-up strategy, linearly increasing the learning rate from 0 up to 0.024 in the first 40 epochs.
Data augmentation was solely done through horizontal flipping, cautiously changing the tooth classes to their new corresponding numbers (right-sided teeth turned into left-sided teeth and vice-versa).
Finally, we mention that the mask branches are class agnostic, \textit{i.e.}, it only segments the object from the background.
Table \ref{tab:benchmark-results} summarizes the benchmark results (with the scores in green representing the highest one, while the smallest one are in red).
The winner architecture was HTC, which also had the best values on all considered metrics, except on segmentation AP50 by a tiny margin.
The HTC's final scores were also substantial, confirming it as a trustworthy option for our HITL scheme.
We used the benchmark's resulting HTC neural network to start the labeling of new radiographs.

\begin{table}[t]
\centering
\caption{Summary of the benchmark results. The HTC architecture with DCN backboned by a ResNeXt-101-64x4d was the winner architecture and the used model for our HITL scheme.}
\label{tab:benchmark-results}
\resizebox{\textwidth}{!}{%
\begin{tabular}{|l|c|c|l|ccc|ccc|}
\hline
\rowcolor[HTML]{EFEFEF} 
\cellcolor[HTML]{EFEFEF} &
  \cellcolor[HTML]{EFEFEF} &
  \cellcolor[HTML]{EFEFEF} &
  \multicolumn{1}{c|}{\cellcolor[HTML]{EFEFEF}} &
  \multicolumn{3}{c|}{\cellcolor[HTML]{EFEFEF}Detection} &
  \multicolumn{3}{c|}{\cellcolor[HTML]{EFEFEF}Segmentation} \\ \cline{5-10} 
\rowcolor[HTML]{EFEFEF} 
\multirow{-2}{*}{\cellcolor[HTML]{EFEFEF}Architecture} &
  \multirow{-2}{*}{\cellcolor[HTML]{EFEFEF}Backbone} &
  \multirow{-2}{*}{\cellcolor[HTML]{EFEFEF}Head} &
  \multicolumn{1}{c|}{\multirow{-2}{*}{\cellcolor[HTML]{EFEFEF}DCN}} &
  \multicolumn{1}{c|}{\cellcolor[HTML]{EFEFEF}AP75} &
  \multicolumn{1}{c|}{\cellcolor[HTML]{EFEFEF}AP50} &
  mAP &
  \multicolumn{1}{c|}{\cellcolor[HTML]{EFEFEF}AP75} &
  \multicolumn{1}{c|}{\cellcolor[HTML]{EFEFEF}AP50} &
  mAP \\ \hline
\rowcolor[HTML]{EFEFEF} 
HTC &
  X-101-64x4d-FPN &
  HTC &
  \multicolumn{1}{c|}{\cellcolor[HTML]{EFEFEF}\checkmark} &
  \multicolumn{1}{c|}{\cellcolor[HTML]{EFEFEF}{\color[HTML]{34A853} \textbf{0.913}}} &
  \multicolumn{1}{c|}{\cellcolor[HTML]{EFEFEF}{\color[HTML]{34A853} \textbf{0.983}}} &
  {\color[HTML]{34A853} \textbf{0.795}} &
  \multicolumn{1}{c|}{\cellcolor[HTML]{EFEFEF}{\color[HTML]{34A853} \textbf{0.958}}} &
  \multicolumn{1}{c|}{\cellcolor[HTML]{EFEFEF}0.983} &
  {\color[HTML]{34A853} \textbf{0.802}} \\ \cline{1-4}
DetectoRS &
  ResNet-50 &
  HTC &
   &
  \multicolumn{1}{c|}{0.909} &
  \multicolumn{1}{c|}{0.982} &
  0.777 &
  \multicolumn{1}{c|}{0.930} &
  \multicolumn{1}{c|}{{\color[HTML]{34A853} \textbf{0.984}}} &
  0.780 \\ \cline{1-4}
\rowcolor[HTML]{EFEFEF} 
ResNeSt Cascade R-CNN &
  S-101-FPN &
  Cascade &
   &
  \multicolumn{1}{c|}{\cellcolor[HTML]{EFEFEF}0.871} &
  \multicolumn{1}{c|}{\cellcolor[HTML]{EFEFEF}{\color[HTML]{EA4335} \textbf{0.917}}} &
  {\color[HTML]{EA4335} \textbf{0.748}} &
  \multicolumn{1}{c|}{\cellcolor[HTML]{EFEFEF}{\color[HTML]{EA4335} \textbf{0.898}}} &
  \multicolumn{1}{c|}{\cellcolor[HTML]{EFEFEF}{\color[HTML]{EA4335} \textbf{0.918}}} &
  {\color[HTML]{EA4335} \textbf{0.745}} \\ \cline{1-4}
Cascade R-CNN with DCN &
  X-101-32x4d-FPN &
  Cascade &
  \multicolumn{1}{c|}{\checkmark} &
  \multicolumn{1}{c|}{0.896} &
  \multicolumn{1}{c|}{0.972} &
  0.771 &
  \multicolumn{1}{c|}{0.922} &
  \multicolumn{1}{c|}{0.972} &
  0.766 \\ \cline{1-4}
\rowcolor[HTML]{EFEFEF} 
ResNeSt Mask R-CNN &
  S-101-FPN &
  FCN &
   &
  \multicolumn{1}{c|}{\cellcolor[HTML]{EFEFEF}0.873} &
  \multicolumn{1}{c|}{\cellcolor[HTML]{EFEFEF}0.931} &
  0.753 &
  \multicolumn{1}{c|}{\cellcolor[HTML]{EFEFEF}0.913} &
  \multicolumn{1}{c|}{\cellcolor[HTML]{EFEFEF}0.931} &
  0.755 \\ \cline{1-4}
Cascade R-CNN &
  X-101-64x4d-FPN &
  Cascade &
   &
  \multicolumn{1}{c|}{0.901} &
  \multicolumn{1}{c|}{0.982} &
  0.763 &
  \multicolumn{1}{c|}{0.939} &
  \multicolumn{1}{c|}{0.982} &
  0.768 \\ \cline{1-4}
\rowcolor[HTML]{EFEFEF} 
Mask R-CNN &
  X-101-64x4d-FPN &
  FCN &
   &
  \multicolumn{1}{c|}{\cellcolor[HTML]{EFEFEF}{\color[HTML]{EA4335} \textbf{0.848}}} &
  \multicolumn{1}{c|}{\cellcolor[HTML]{EFEFEF}0.969} &
  0.752 &
  \multicolumn{1}{c|}{\cellcolor[HTML]{EFEFEF}0.916} &
  \multicolumn{1}{c|}{\cellcolor[HTML]{EFEFEF}0.978} &
  0.758 \\ \hline
\end{tabular}%
}
\end{table}

\subsection{HITL labeling}
\label{section:hitl-labeling}

\begin{figure}[t]
    \centering
    \begin{subfigure}[b]{0.4\textwidth}
        \centering
        \includegraphics[width=\textwidth]{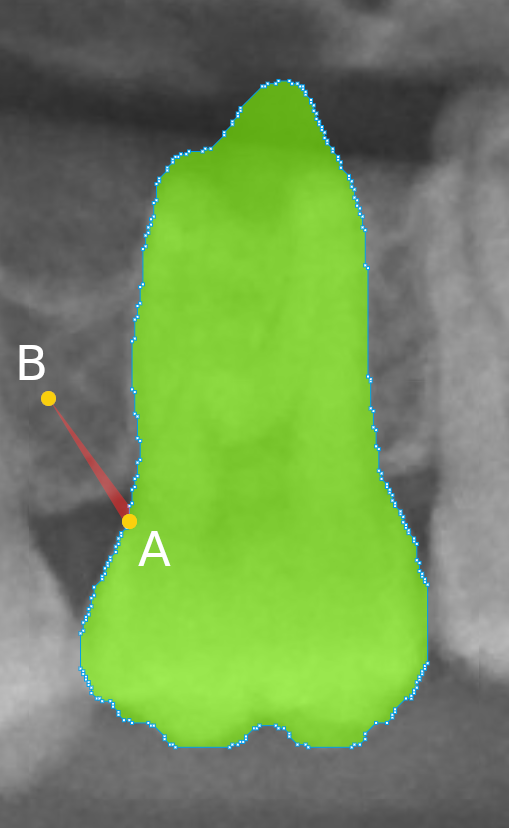}
        \caption{Original code.}
        \label{fig:original-code}
    \end{subfigure}
    \begin{subfigure}[b]{0.4\textwidth}
        \centering
        \includegraphics[width=\textwidth]{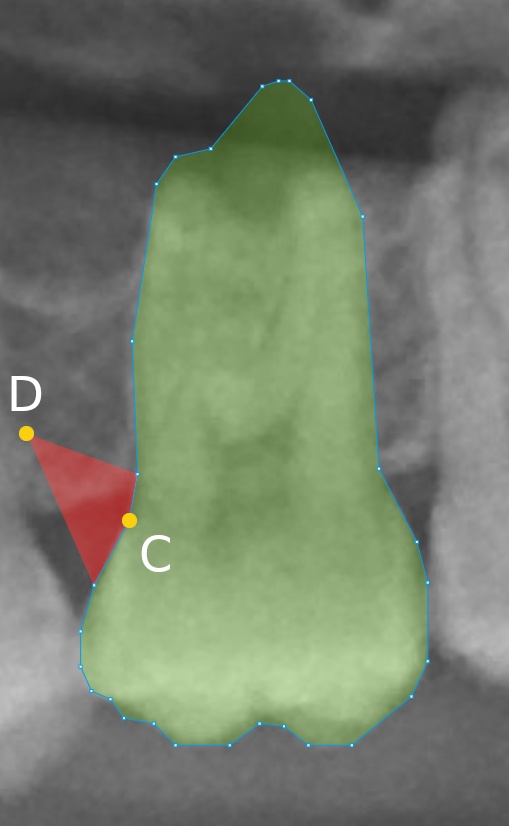}
        \caption{Changed code.}
        \label{fig:changed-code}
    \end{subfigure}
    \caption{Illustration of the software visualization due to the code changes. The red area evinces the higher impact in the annotation when on a point shift. We reduced the shape opacities, easing the annotation verification, and lowered the number of control points. (a) Visualization of a tooth annotation with the original code and no tolerance in the Ramer–Douglas–Peucker algorithm and the impact on the annotation when point A moves to point B. (b) Visualization of a tooth annotation with the changed code and the impact on the annotation when point C moves to point D.}
    \label{fig:point-simplification}
\end{figure}

The HITL-based labeling started with the predictions of the HTC neural network. We labeled 450 radiographs in the first HITL iteration, as indicated in Figure \ref{fig:hitl-setup}.
This iteration was considered experimental, as the annotators had not previously verified annotations from model predictions.
Indeed, it quickly became notorious that manual image labeling is quite different from labeling verification.
When labeling a radiograph from scratch, the annotator may promptly detect or localize the teeth and segment their instances using the annotator software mechanisms such as the polygon or brush tools.
In the COCO Annotator software, the resulting area is filled with a colored layer to distinguish the already segmented objects from the others.
On the other hand, when working on verifying neural network predictions, the human annotators must visually inspect the results and quickly confirm or correct the provisional labels.
For that, the annotators can benefit from any software annotation tools, but in our case, they most frequently used the polygon point drag-and-drop feature.
Two issues arise from this: (i) the filled segmented areas obstruct the instances, hampering the verification; (ii) the large number of points per segmentation slows down and hardens the corrections because point shift has less impact on the annotation.
We mitigated these issues by changing the software source code, reducing the shape opacity, and lowering the number of control points through the Ramer–Douglas–Peucker algorithm with a tolerance of 2 pixels (\citeauthor{douglas1973algorithms}, \citeyear{douglas1973algorithms}).
Figure \ref{fig:point-simplification} illustrates these modifications, evincing the new higher impact of point shift.
Furthermore, we added a keyboard shortcut to toggle the annotation visualization, which was very helpful for the annotators.

\begin{figure}[t]
    \centering
    \begin{subfigure}[b]{0.49\textwidth}
        \centering
        \includegraphics[width=\textwidth]{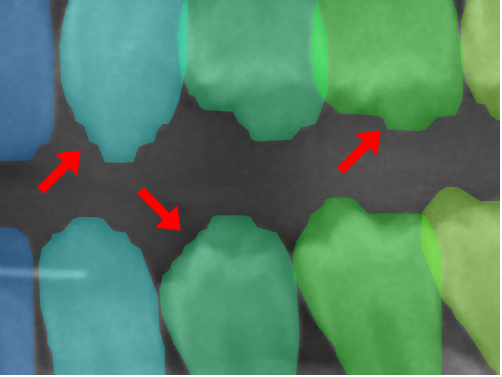}
        \caption{Serrated masks on tooth crowns.}
        \label{subfig:crowns}
    \end{subfigure}
    \hfill
    \begin{subfigure}[b]{0.49\textwidth}
        \centering
        \includegraphics[width=\textwidth]{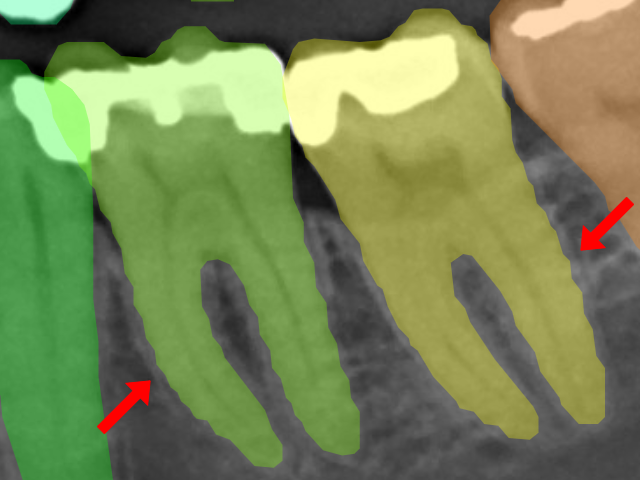}
        \caption{Serrated masks on lower molars.}
        \label{subfig:molars}
    \end{subfigure}
    \caption{Samples of the serrated pattern due to the network's low-resolution mask predictions. This pattern frequently occurred on (a) tooth crowns and (b) lower molars, especially the former. The red arrows point to some spots with those patterns.}
    \label{fig:low-mask-samples}
\end{figure}

We defined some correction criteria based on our observations during the labeling verification of the first HITL iteration.
It was evident that the network predictions were outstanding, yet they were usually worse than manual annotations.
This worse performance was mainly due to delicate details that could be polished such as the serrated segmentations originated from the network's low-resolution masks.
Figure \ref{fig:low-mask-samples} shows samples of the serrated patterns on tooth crowns and on lower molars, which were highly frequent, especially on the former.
For many applications, such as tooth detection and numbering, these tiny mistakes can be overlooked.
However, we decided not to ignore those errors, as we want our data set to be general-purpose.
In sum, the labels after correction should be as similar as possible to the manual labels.
This determination slowed down the verification procedure significantly because our annotators had to make many tiny adjustments. 
With this main criterion defined, we proceeded with the other three iterations, reaching in the end 3,150 HITL labeled radiographs.

\section{Evaluation of the HITL results}

\begin{table}[t]
\centering
\caption{Results of the trained neural networks in our HITL system on their corresponding validation data sets. The validation data sets comprise 10\% of the available data at their HITL iteration. We highlight the best (green) and worst (red) results per metric.}
\label{tab:hitl-validation-results}
\begin{tabular}{|c|ccc|ccc|}
\hline
\rowcolor[HTML]{EFEFEF} 
\cellcolor[HTML]{EFEFEF} &
  \multicolumn{3}{c|}{\cellcolor[HTML]{EFEFEF}Detection} &
  \multicolumn{3}{c|}{\cellcolor[HTML]{EFEFEF}Segmentation} \\ \cline{2-7} 
\rowcolor[HTML]{EFEFEF} 
\multirow{-2}{*}{\cellcolor[HTML]{EFEFEF}\begin{tabular}[c]{@{}c@{}}Neural\\ Networks\end{tabular}} &
  \multicolumn{1}{c|}{\cellcolor[HTML]{EFEFEF}AP50} &
  \multicolumn{1}{c|}{\cellcolor[HTML]{EFEFEF}AP75} &
  mAP &
  \multicolumn{1}{c|}{\cellcolor[HTML]{EFEFEF}AP50} &
  \multicolumn{1}{c|}{\cellcolor[HTML]{EFEFEF}AP75} &
  mAP \\ \hline
HTC 1 &
  \multicolumn{1}{c|}{{\color[HTML]{EA4335} \textbf{98.3}}} &
  \multicolumn{1}{c|}{{\color[HTML]{EA4335} \textbf{91.3}}} &
  {\color[HTML]{EA4335} \textbf{79.5}} &
  \multicolumn{1}{c|}{{\color[HTML]{EA4335} \textbf{98.3}}} &
  \multicolumn{1}{c|}{{\color[HTML]{EA4335} \textbf{95.8}}} &
  {\color[HTML]{EA4335} \textbf{80.2}} \\ \cline{1-1}
\rowcolor[HTML]{EFEFEF} 
HTC 2 &
  \multicolumn{1}{c|}{\cellcolor[HTML]{EFEFEF}98.7} &
  \multicolumn{1}{c|}{\cellcolor[HTML]{EFEFEF}94.8} &
  81.6 &
  \multicolumn{1}{c|}{\cellcolor[HTML]{EFEFEF}98.7} &
  \multicolumn{1}{c|}{\cellcolor[HTML]{EFEFEF}96.7} &
  82.1 \\ \cline{1-1}
HTC 3 &
  \multicolumn{1}{c|}{{\color[HTML]{34A853} \textbf{98.9}}} &
  \multicolumn{1}{c|}{{\color[HTML]{34A853} \textbf{97.1}}} &
  83.6 &
  \multicolumn{1}{c|}{{\color[HTML]{34A853} \textbf{98.9}}} &
  \multicolumn{1}{c|}{97.1} &
  83.6 \\ \cline{1-1}
\rowcolor[HTML]{EFEFEF} 
HTC 4 &
  \multicolumn{1}{c|}{\cellcolor[HTML]{EFEFEF}{\color[HTML]{34A853} \textbf{98.9}}} &
  \multicolumn{1}{c|}{\cellcolor[HTML]{EFEFEF}96.6} &
  {\color[HTML]{34A853} \textbf{86}} &
  \multicolumn{1}{c|}{\cellcolor[HTML]{EFEFEF}{\color[HTML]{34A853} \textbf{98.9}}} &
  \multicolumn{1}{c|}{\cellcolor[HTML]{EFEFEF}{\color[HTML]{34A853} \textbf{97.7}}} &
  {\color[HTML]{34A853} \textbf{85.9}} \\ \hline
\end{tabular}
\end{table}

The ultimate goal of our work was to create a labeled data set to boost research on dental panoramic radiographs.
Under this perspective, the outcomes of the HITL procedure sufficed for our purpose, dispensing to report the performance of the trained models on a test set.
However, we expect the deep learning community to heavily use our data set and increasingly employ the HITL concept to speed up the annotation process.
Therefore, we performed detailed analyses on the HITL outcomes, including the evaluation of the trained networks on a separate manually labeled test data set.
We aimed through these analyses to measure the HITL benefits and identify the main bottlenecks for better results.

\subsection{Model results on validation data}
\label{section:results-on-validation-sets}

Table \ref{tab:hitl-validation-results} synthesizes the detection and segmentation metrics (AP50, AP75, and mAP) attained by the trained neural networks in our HITL system, highlighting the best (green) and worst (red) results per metric.
These metrics come from the best networks according to the segmentation mAP over the validation data sets, which comprised 10\% of the available data at each HITL iteration.
We call these networks HTC 1, HTC 2, HTC 3, and HTC 4.
The number in their names corresponds to the iteration in which the network was trained.

When looking at the results of Table \ref{tab:hitl-validation-results}, we perceive an unmistakable increasing trend on the considered metrics, especially on the mAP ones, which contain the primary metric (mAP for segmentation).
The increasing trend exists in the other looser metrics (AP50 and AP75) but is less pronounced.
This difference was no surprise, as the selection of the network weights was according to the segmentation mAP, and the AP50 and AP75 values were already pretty high on the first HITL iteration (not much room for improvement).

\subsection{Model results on HITL data}
\label{section:results-on-hitl-data}

The HITL labeled data is an alternative to the validation data sets for model evaluation.
In this case, we evaluate the model performances on the verified annotations from the model predictions.
The main advantage here is that we do model assessment in unseen and large data.
We performed this analysis using the threshold values computed with the procedure described in Section \ref{section:hitl-labeling}.
Table \ref{tab:hitl-verified-data-results} synthesizes the results of HTC 1, 2, and 3 on, respectively, 450, 900, and 1800 radiographs labeled from their corresponding predictions, also highlighting the best and worst results (we did not assess HTC 4 as no labels came from its predictions).
All metrics increased at each iteration, but the most significant performance boost came from HTC 1 to HTC 2, when there was still significant room for improvement.
The mAPs of HTC 3 were the best and surpassed the 80 points on both the detection and segmentation tasks.

\begin{table}[t]
\centering
\caption{Results of HTC 1, 2, and 3 on the verified labeled from their predictions over 450, 900, and 1800 images, respectively.
We highlight the best (green) and worst (red) results per metric.}
\label{tab:hitl-verified-data-results}
\begin{tabular}{|c|ccc|ccc|}
\hline
\rowcolor[HTML]{EFEFEF} 
\cellcolor[HTML]{EFEFEF} &
  \multicolumn{3}{c|}{\cellcolor[HTML]{EFEFEF}Detection} &
  \multicolumn{3}{c|}{\cellcolor[HTML]{EFEFEF}Segmentation} \\ \cline{2-7} 
\rowcolor[HTML]{EFEFEF} 
\multirow{-2}{*}{\cellcolor[HTML]{EFEFEF}\begin{tabular}[c]{@{}c@{}}Neural\\ Networks\end{tabular}} &
  \multicolumn{1}{c|}{\cellcolor[HTML]{EFEFEF}AP50} &
  \multicolumn{1}{c|}{\cellcolor[HTML]{EFEFEF}AP75} &
  mAP &
  \multicolumn{1}{c|}{\cellcolor[HTML]{EFEFEF}AP50} &
  \multicolumn{1}{c|}{\cellcolor[HTML]{EFEFEF}AP75} &
  mAP \\ \hline
HTC 1 &
  \multicolumn{1}{c|}{{\color[HTML]{EA4335} \textbf{82.0}}} &
  \multicolumn{1}{c|}{{\color[HTML]{EA4335} \textbf{79.2}}} &
  {\color[HTML]{EA4335} \textbf{71.2}} &
  \multicolumn{1}{c|}{{\color[HTML]{EA4335} \textbf{82.0}}} &
  \multicolumn{1}{c|}{{\color[HTML]{EA4335} \textbf{80.3}}} &
  {\color[HTML]{EA4335} \textbf{74.0}} \\ \cline{1-1}
\rowcolor[HTML]{EFEFEF} 
HTC 2 &
  \multicolumn{1}{c|}{\cellcolor[HTML]{EFEFEF}88.4} &
  \multicolumn{1}{c|}{\cellcolor[HTML]{EFEFEF}87.5} &
  79.6 &
  \multicolumn{1}{c|}{\cellcolor[HTML]{EFEFEF}88.4} &
  \multicolumn{1}{c|}{\cellcolor[HTML]{EFEFEF}87.6} &
  82.1 \\ \cline{1-1}
HTC 3 &
  \multicolumn{1}{c|}{{\color[HTML]{34A853} \textbf{89.4}}} &
  \multicolumn{1}{c|}{{\color[HTML]{34A853} \textbf{88.3}}} &
  {\color[HTML]{34A853} \textbf{80.9}} &
  \multicolumn{1}{c|}{{\color[HTML]{34A853} \textbf{89.4}}} &
  \multicolumn{1}{c|}{{\color[HTML]{34A853} \textbf{88.7}}} &
  {\color[HTML]{34A853} \textbf{82.7}} \\ \hline
\end{tabular}
\end{table}

Using the HITL labeled data as test data mitigated the problem of the biased estimation of the network results and the computed metrics revealed consistent results.
However, some issues persisted: we evaluated the networks on distinct images with different radiograph category proportions and disregarded HTC 4.
For those reasons, it proved imperative to label a separate set of images for a consistent comparison.
For that, we assessed the networks on 400 images (40 for each radiograph category), which we manually labeled from scratch and comprised our test data set.

\subsection{Model results on test data}
\label{section:results-on-test-data}

Besides a consistent comparison, our test data set allows unbiased model assessment, as we manually labeled 40 images per radiograph category exclusively for model evaluation.
Table \ref{tab:networks-on-test} synthesizes the results of each trained HTC network over the test data set accordingly to the detection and segmentation AP50, AP75, and mAP metrics.
One can observe that all segmentation metrics increased at each HITL cycle, being a favorable indication for the HITL results.
The detection metrics also display a prominent increasing tendency, but they may oscillate slightly.
These aggregate results give no insights on the specifics of the network performances.
In order to solve that, we analyzed the segmentation mAP per dentition and tooth type.

\begin{table}[t]
\centering
\caption{Performance metrics of each trained neural network in our HITL system on the manually annotated test data set. The test data set comprised 400 images (40 per radiograph category). We highlight the best (green) and worst (red) results per metric.}
\label{tab:networks-on-test}
\begin{tabular}{|c|ccc|ccc|}
\hline
\rowcolor[HTML]{EFEFEF} 
\cellcolor[HTML]{EFEFEF} &
  \multicolumn{3}{c|}{\cellcolor[HTML]{EFEFEF}Detection} &
  \multicolumn{3}{c|}{\cellcolor[HTML]{EFEFEF}Segmentation} \\ \cline{2-7} 
\rowcolor[HTML]{EFEFEF} 
\multirow{-2}{*}{\cellcolor[HTML]{EFEFEF}\begin{tabular}[c]{@{}c@{}}Neural\\ Networks\end{tabular}} &
  \multicolumn{1}{c|}{\cellcolor[HTML]{EFEFEF}AP50} &
  \multicolumn{1}{c|}{\cellcolor[HTML]{EFEFEF}AP75} &
  mAP &
  \multicolumn{1}{c|}{\cellcolor[HTML]{EFEFEF}AP50} &
  \multicolumn{1}{c|}{\cellcolor[HTML]{EFEFEF}AP75} &
  mAP \\ \hline
HTC 1 &
  \multicolumn{1}{c|}{{\color[HTML]{EA4335} \textbf{91.9}}} &
  \multicolumn{1}{c|}{{\color[HTML]{EA4335} \textbf{83.4}}} &
  \cellcolor[HTML]{FFFFFF}{\color[HTML]{EA4335} \textbf{72.0}} &
  \multicolumn{1}{c|}{{\color[HTML]{EA4335} \textbf{92.2}}} &
  \multicolumn{1}{c|}{{\color[HTML]{EA4335} \textbf{87.2}}} &
  {\color[HTML]{EA4335} \textbf{72.0}} \\ \cline{1-1}
\rowcolor[HTML]{EFEFEF} 
HTC 2 &
  \multicolumn{1}{c|}{\cellcolor[HTML]{EFEFEF}95.4} &
  \multicolumn{1}{c|}{\cellcolor[HTML]{EFEFEF}87.8} &
  75.5 &
  \multicolumn{1}{c|}{\cellcolor[HTML]{EFEFEF}95.7} &
  \multicolumn{1}{c|}{\cellcolor[HTML]{EFEFEF}89.5} &
  74.6 \\ \cline{1-1}
HTC 3 &
  \multicolumn{1}{c|}{97.0} &
  \multicolumn{1}{c|}{{\color[HTML]{34A853} \textbf{88.0}}} &
  75.4 &
  \multicolumn{1}{c|}{97.6} &
  \multicolumn{1}{c|}{\cellcolor[HTML]{FFFFFF}90.3} &
  75.6 \\ \cline{1-1}
\rowcolor[HTML]{EFEFEF} 
HTC 4 &
  \multicolumn{1}{c|}{\cellcolor[HTML]{EFEFEF}{\color[HTML]{34A853} \textbf{98.4}}} &
  \multicolumn{1}{c|}{\cellcolor[HTML]{EFEFEF}87.4} &
  {\color[HTML]{34A853} \textbf{76.0}} &
  \multicolumn{1}{c|}{\cellcolor[HTML]{EFEFEF}{\color[HTML]{34A853} \textbf{98.9}}} &
  \multicolumn{1}{c|}{\cellcolor[HTML]{EFEFEF}{\color[HTML]{34A853} \textbf{91.8}}} &
  {\color[HTML]{34A853} \textbf{77.4}} \\ \hline
\end{tabular}
\end{table}

Table \ref{tab:perm-and-deciduous-map} split the segmentation metrics into the dentition types: permanent and deciduous.
The highest (green) and smallest (red) values per metric indicate that the best predictions are from HTC 4, while the worst are from HTC 1.
These metrics demonstrate small but consistent increasing performances over the HITL iterations on permanent dentitions.
On the other hand, the segmentation results on deciduous teeth improved significantly: at least 15\% on all metrics.
The segmentation mAP increased 12.9 points, which represents a 23.0\% gain.
This improvement is due to the initial lack of training data increment (the deciduous teeth constitute approximately 3\% of the training instances).
The rare occurrences of deciduous teeth, especially the central and lower lateral incisors, hindered the first trained networks from generalizing on those tooth types.

\begin{table}[t]
\centering
\caption{Segmentation results on the test data set according to the dentition tooth types: Permanent and deciduous. We highlight the best (green) and worst (red) results per metric. The metrics over the deciduous teeth were worse but improved significantly over the HITL iterations.}
\label{tab:perm-and-deciduous-map}
\begin{tabular}{|c|ccc|ccc|}
\hline
\rowcolor[HTML]{EFEFEF} 
\cellcolor[HTML]{EFEFEF} &
  \multicolumn{3}{c|}{\cellcolor[HTML]{EFEFEF}Permanent} &
  \multicolumn{3}{c|}{\cellcolor[HTML]{EFEFEF}Deciduous} \\ \cline{2-7} 
\rowcolor[HTML]{EFEFEF} 
\multirow{-2}{*}{\cellcolor[HTML]{EFEFEF}\begin{tabular}[c]{@{}c@{}}Neural\\ Network\end{tabular}} &
  \multicolumn{1}{c|}{\cellcolor[HTML]{EFEFEF}AP50} &
  \multicolumn{1}{c|}{\cellcolor[HTML]{EFEFEF}AP75} &
  mAP &
  \multicolumn{1}{c|}{\cellcolor[HTML]{EFEFEF}AP50} &
  \multicolumn{1}{c|}{\cellcolor[HTML]{EFEFEF}AP75} &
  mAP \\ \hline
HTC 1 &
  \multicolumn{1}{c|}{{\color[HTML]{EA4335} \textbf{99.0}}} &
  \multicolumn{1}{c|}{{\color[HTML]{EA4335} \textbf{97.0}}} &
  \cellcolor[HTML]{FFFFFF}{\color[HTML]{EA4335} \textbf{82.0}} &
  \multicolumn{1}{c|}{{\color[HTML]{EA4335} \textbf{81.5}}} &
  \multicolumn{1}{c|}{{\color[HTML]{EA4335} \textbf{71.4}}} &
  {\color[HTML]{EA4335} \textbf{56.1}} \\ \cline{1-1}
\rowcolor[HTML]{EFEFEF} 
HTC 2 &
  \multicolumn{1}{c|}{\cellcolor[HTML]{EFEFEF}{\color[HTML]{EA4335} \textbf{99.0}}} &
  \multicolumn{1}{c|}{\cellcolor[HTML]{EFEFEF}97.3} &
  82.3 &
  \multicolumn{1}{c|}{\cellcolor[HTML]{EFEFEF}90.4} &
  \multicolumn{1}{c|}{\cellcolor[HTML]{EFEFEF}77.0} &
  62.3 \\ \cline{1-1}
HTC 3 &
  \multicolumn{1}{c|}{{\color[HTML]{34A853} \textbf{99.1}}} &
  \multicolumn{1}{c|}{{\color[HTML]{34A853} \textbf{97.5}}} &
  82.4 &
  \multicolumn{1}{c|}{95.3} &
  \multicolumn{1}{c|}{\cellcolor[HTML]{FFFFFF}78.7} &
  64.7 \\ \cline{1-1}
\rowcolor[HTML]{EFEFEF} 
HTC 4 &
  \multicolumn{1}{c|}{\cellcolor[HTML]{EFEFEF}{\color[HTML]{34A853} \textbf{99.1}}} &
  \multicolumn{1}{c|}{\cellcolor[HTML]{EFEFEF}{\color[HTML]{34A853} \textbf{97.6}}} &
  {\color[HTML]{34A853} \textbf{82.7}} &
  \multicolumn{1}{c|}{\cellcolor[HTML]{EFEFEF}{\color[HTML]{34A853} \textbf{98.5}}} &
  \multicolumn{1}{c|}{\cellcolor[HTML]{EFEFEF}{\color[HTML]{34A853} \textbf{82.7}}} &
  {\color[HTML]{34A853} \textbf{69.0}} \\ \hline
\end{tabular}
\end{table}

Finally, we broke the segmentation mAP by dentition and tooth type in Tables \ref{tab:map-permanent-teeth} (permanent teeth) and \ref{tab:map-deciduous-teeth} (deciduous teeth), highlighting the best (green) and worst (red) results of the networks per tooth type.
Table \ref{tab:map-deciduous-teeth} also brings the number of instances presented in the training data sets to illustrate the initial lack of training data.
No lower central incisor was present in the first training iteration (HTC 1), and only 14 were present in the last (HTC 4).
The additional data on those less frequent tooth types resulted in significantly higher metric values.
Table \ref{tab:map-permanent-teeth} unveils that, on the permanent teeth, the HITL was more beneficial on the segmentation of the upper teeth than the lower ones.
HTC 4 performed better on permanent lower incisors and permanent upper teeth than HTC 1.
According to our annotators, the upper teeth, particularly the premolars and molars, are harder to segment.
We consider this fact, along with the improved metrics on those tooth types propitiated by the HITL scheme, to subsidize the use of deep learning-based assist tools to aid in challenging cases.
In contrast, the metrics on permanent lower premolars and molars stagnated or oscillated a bit.
The metrics on those teeth, which are large and straightforward to segment teeth, were already pretty high on the first iteration, resulting in less room for improvement.

\begin{table}[t]
\centering
\caption{The mAP results on the test data set per permanent tooth type. We highlight the best (green) and worst (red) results per tooth. The HITL benefited more the metrics over the more challenging to segment teeth, such as the upper premolars and molars.}
\label{tab:map-permanent-teeth}
\resizebox{\textwidth}{!}{%
\begin{tabular}{|cccccccccc|}
\hline
\rowcolor[HTML]{EFEFEF} 
\multicolumn{1}{|c|}{\cellcolor[HTML]{EFEFEF}} &
  \multicolumn{1}{c|}{\cellcolor[HTML]{EFEFEF}} &
  \multicolumn{2}{c|}{\cellcolor[HTML]{EFEFEF}Incisors} &
  \multicolumn{1}{c|}{\cellcolor[HTML]{EFEFEF}} &
  \multicolumn{2}{c|}{\cellcolor[HTML]{EFEFEF}Premolars} &
  \multicolumn{3}{c|}{\cellcolor[HTML]{EFEFEF}Molars} \\ \cline{3-4} \cline{6-10} 
\rowcolor[HTML]{EFEFEF} 
\multicolumn{1}{|c|}{\multirow{-2}{*}{\cellcolor[HTML]{EFEFEF}\begin{tabular}[c]{@{}c@{}}Dental\\ Arch\end{tabular}}} &
  \multicolumn{1}{c|}{\multirow{-2}{*}{\cellcolor[HTML]{EFEFEF}\begin{tabular}[c]{@{}c@{}}Neural\\ Network\end{tabular}}} &
  \multicolumn{1}{c|}{\cellcolor[HTML]{EFEFEF}Central} &
  \multicolumn{1}{c|}{\cellcolor[HTML]{EFEFEF}Lateral} &
  \multicolumn{1}{c|}{\multirow{-2}{*}{\cellcolor[HTML]{EFEFEF}Canines}} &
  \multicolumn{1}{c|}{\cellcolor[HTML]{EFEFEF}1st} &
  \multicolumn{1}{c|}{\cellcolor[HTML]{EFEFEF}2nd} &
  \multicolumn{1}{c|}{\cellcolor[HTML]{EFEFEF}1st} &
  \multicolumn{1}{c|}{\cellcolor[HTML]{EFEFEF}2nd} &
  3rd \\ \hline
\multicolumn{1}{|c|}{\cellcolor[HTML]{EFEFEF}} &
  \multicolumn{1}{c|}{HTC 1} &
  \multicolumn{1}{c|}{{\color[HTML]{EA4335} \textbf{85.1}}} &
  \multicolumn{1}{c|}{{\color[HTML]{EA4335} \textbf{83.8}}} &
  \multicolumn{1}{c|}{{\color[HTML]{EA4335} \textbf{84.0}}} &
  \multicolumn{1}{c|}{{\color[HTML]{EA4335} \textbf{72.9}}} &
  \multicolumn{1}{c|}{{\color[HTML]{EA4335} \textbf{79.9}}} &
  \multicolumn{1}{c|}{{\color[HTML]{EA4335} \textbf{78.4}}} &
  \multicolumn{1}{c|}{{\color[HTML]{EA4335} \textbf{80.7}}} &
  {\color[HTML]{EA4335} \textbf{77.2}} \\ \cline{2-2}
\rowcolor[HTML]{EFEFEF} 
\multicolumn{1}{|c|}{\cellcolor[HTML]{EFEFEF}} &
  \multicolumn{1}{c|}{\cellcolor[HTML]{EFEFEF}HTC 2} &
  \multicolumn{1}{c|}{\cellcolor[HTML]{EFEFEF}85.8} &
  \multicolumn{1}{c|}{\cellcolor[HTML]{EFEFEF}83.9} &
  \multicolumn{1}{c|}{\cellcolor[HTML]{EFEFEF}84.4} &
  \multicolumn{1}{c|}{\cellcolor[HTML]{EFEFEF}73.4} &
  \multicolumn{1}{c|}{\cellcolor[HTML]{EFEFEF}{\color[HTML]{EA4335} \textbf{79.9}}} &
  \multicolumn{1}{c|}{\cellcolor[HTML]{EFEFEF}80.0} &
  \multicolumn{1}{c|}{\cellcolor[HTML]{EFEFEF}80.8} &
  77.7 \\ \cline{2-2}
\multicolumn{1}{|c|}{\cellcolor[HTML]{EFEFEF}} &
  \multicolumn{1}{c|}{HTC 3} &
  \multicolumn{1}{c|}{85.8} &
  \multicolumn{1}{c|}{84.0} &
  \multicolumn{1}{c|}{{\color[HTML]{34A853} \textbf{84.8}}} &
  \multicolumn{1}{c|}{73.4} &
  \multicolumn{1}{c|}{81.0} &
  \multicolumn{1}{c|}{{\color[HTML]{34A853} \textbf{80.1}}} &
  \multicolumn{1}{c|}{81.1} &
  77.8 \\ \cline{2-2}
\rowcolor[HTML]{EFEFEF} 
\multicolumn{1}{|c|}{\multirow{-4}{*}{\cellcolor[HTML]{EFEFEF}Upper}} &
  \multicolumn{1}{c|}{\cellcolor[HTML]{EFEFEF}HTC 4} &
  \multicolumn{1}{c|}{\cellcolor[HTML]{EFEFEF}{\color[HTML]{34A853} \textbf{85.9}}} &
  \multicolumn{1}{c|}{\cellcolor[HTML]{EFEFEF}{\color[HTML]{34A853} \textbf{84.4}}} &
  \multicolumn{1}{c|}{\cellcolor[HTML]{EFEFEF}84.5} &
  \multicolumn{1}{c|}{\cellcolor[HTML]{EFEFEF}{\color[HTML]{34A853} \textbf{74.7}}} &
  \multicolumn{1}{c|}{\cellcolor[HTML]{EFEFEF}{\color[HTML]{34A853} \textbf{81.1}}} &
  \multicolumn{1}{c|}{\cellcolor[HTML]{EFEFEF}79.9} &
  \multicolumn{1}{c|}{\cellcolor[HTML]{EFEFEF}{\color[HTML]{34A853} \textbf{81.6}}} &
  {\color[HTML]{34A853} \textbf{78.0}} \\ \hline
\multicolumn{1}{|c|}{\cellcolor[HTML]{EFEFEF}} &
  \multicolumn{1}{c|}{HTC 1} &
  \multicolumn{1}{c|}{{\color[HTML]{EA4335} \textbf{79.0}}} &
  \multicolumn{1}{c|}{{\color[HTML]{EA4335} \textbf{80.8}}} &
  \multicolumn{1}{c|}{{\color[HTML]{EA4335} \textbf{85.3}}} &
  \multicolumn{1}{c|}{{\color[HTML]{EA4335} \textbf{84.7}}} &
  \multicolumn{1}{c|}{{\color[HTML]{34A853} \textbf{87.3}}} &
  \multicolumn{1}{c|}{{\color[HTML]{34A853} \textbf{85.1}}} &
  \multicolumn{1}{c|}{{\color[HTML]{34A853} \textbf{84.5}}} &
  {\color[HTML]{EA4335} \textbf{82.9}} \\ \cline{2-2}
\rowcolor[HTML]{EFEFEF} 
\multicolumn{1}{|c|}{\cellcolor[HTML]{EFEFEF}} &
  \multicolumn{1}{c|}{\cellcolor[HTML]{EFEFEF}HTC 2} &
  \multicolumn{1}{c|}{\cellcolor[HTML]{EFEFEF}79.6} &
  \multicolumn{1}{c|}{\cellcolor[HTML]{EFEFEF}80.9} &
  \multicolumn{1}{c|}{\cellcolor[HTML]{EFEFEF}{\color[HTML]{34A853} \textbf{85.9}}} &
  \multicolumn{1}{c|}{\cellcolor[HTML]{EFEFEF}{\color[HTML]{34A853} \textbf{85.4}}} &
  \multicolumn{1}{c|}{\cellcolor[HTML]{EFEFEF}87.1} &
  \multicolumn{1}{c|}{\cellcolor[HTML]{EFEFEF}84.4} &
  \multicolumn{1}{c|}{\cellcolor[HTML]{EFEFEF}84.1} &
  83.5 \\ \cline{2-2}
\multicolumn{1}{|c|}{\cellcolor[HTML]{EFEFEF}} &
  \multicolumn{1}{c|}{HTC 3} &
  \multicolumn{1}{c|}{{\color[HTML]{34A853} \textbf{79.9}}} &
  \multicolumn{1}{c|}{81.2} &
  \multicolumn{1}{c|}{85.8} &
  \multicolumn{1}{c|}{85.0} &
  \multicolumn{1}{c|}{{\color[HTML]{EA4335} \textbf{87.0}}} &
  \multicolumn{1}{c|}{{\color[HTML]{EA4335} \textbf{84.0}}} &
  \multicolumn{1}{c|}{{\color[HTML]{EA4335} \textbf{83.8}}} &
  {\color[HTML]{EA4335} \textbf{82.9}} \\ \cline{2-2}
\rowcolor[HTML]{EFEFEF} 
\multicolumn{1}{|c|}{\multirow{-4}{*}{\cellcolor[HTML]{EFEFEF}Lower}} &
  \multicolumn{1}{c|}{\cellcolor[HTML]{EFEFEF}HTC 4} &
  \multicolumn{1}{c|}{\cellcolor[HTML]{EFEFEF}79.6} &
  \multicolumn{1}{c|}{\cellcolor[HTML]{EFEFEF}{\color[HTML]{34A853} \textbf{81.6}}} &
  \multicolumn{1}{c|}{\cellcolor[HTML]{EFEFEF}85.7} &
  \multicolumn{1}{c|}{\cellcolor[HTML]{EFEFEF}85.3} &
  \multicolumn{1}{c|}{{\color[HTML]{34A853} \textbf{87.3}}} &
  \multicolumn{1}{c|}{\cellcolor[HTML]{EFEFEF}84.6} &
  \multicolumn{1}{c|}{\cellcolor[HTML]{EFEFEF}{\color[HTML]{34A853} \textbf{84.5}}} &
  {\color[HTML]{34A853} \textbf{83.8}} \\ \hline
\end{tabular}%
}
\end{table}

\begin{table}[t]
\centering
\caption{The mAP results on test data set and instance count (in parentheses) on training sets per deciduous tooth type. We highlight the best (green) and worst (red) results per metric. The metrics over the deciduous teeth were worse on average but improved significantly over the HITL iterations.}
\label{tab:map-deciduous-teeth}
\resizebox{\textwidth}{!}{%
\begin{tabular}{|c|c|cc|c|cc|}
\hline
\rowcolor[HTML]{EFEFEF} 
\cellcolor[HTML]{EFEFEF} &
  \cellcolor[HTML]{EFEFEF} &
  \multicolumn{2}{c|}{\cellcolor[HTML]{EFEFEF}Incisors} &
  \cellcolor[HTML]{EFEFEF} &
  \multicolumn{2}{c|}{\cellcolor[HTML]{EFEFEF}Molars} \\ \cline{3-4} \cline{6-7} 
\rowcolor[HTML]{EFEFEF} 
\multirow{-2}{*}{\cellcolor[HTML]{EFEFEF}\begin{tabular}[c]{@{}c@{}}Dental\\ Arch\end{tabular}} &
  \multirow{-2}{*}{\cellcolor[HTML]{EFEFEF}\begin{tabular}[c]{@{}c@{}}Neural\\ Network\end{tabular}} &
  \multicolumn{1}{c|}{\cellcolor[HTML]{EFEFEF}Central} &
  Lateral &
  \multirow{-2}{*}{\cellcolor[HTML]{EFEFEF}Canines} &
  \multicolumn{1}{c|}{\cellcolor[HTML]{EFEFEF}1st} &
  2nd \\ \hline
\cellcolor[HTML]{EFEFEF} &
  \cellcolor[HTML]{EFEFEF}HTC 1 &
  \multicolumn{1}{c|}{{\color[HTML]{EA4335} \textbf{35.2 (3)}}} &
  58.8 (22) &
  {\color[HTML]{EA4335} \textbf{64.8 (70)}} &
  \multicolumn{1}{c|}{64.7 (58)} &
  65.5 (78) \\ \cline{2-2}
\rowcolor[HTML]{EFEFEF} 
\cellcolor[HTML]{EFEFEF} &
  HTC 2 &
  \multicolumn{1}{c|}{\cellcolor[HTML]{EFEFEF}64.6 (7)} &
  {\color[HTML]{EA4335} \textbf{55.1 (28)}} &
  64.9 (110) &
  \multicolumn{1}{c|}{\cellcolor[HTML]{EFEFEF}{\color[HTML]{EA4335} \textbf{59.8 (79)}}} &
  {\color[HTML]{EA4335} \textbf{62.8 (117)}} \\ \cline{2-2}
\cellcolor[HTML]{EFEFEF} &
  \cellcolor[HTML]{EFEFEF}HTC 3 &
  \multicolumn{1}{c|}{66.4 (19)} &
  62 (56) &
  {\color[HTML]{34A853} \textbf{65.2 (206)}} &
  \multicolumn{1}{c|}{59.2 (152)} &
  65.9 (225) \\ \cline{2-2}
\rowcolor[HTML]{EFEFEF} 
\multirow{-4}{*}{\cellcolor[HTML]{EFEFEF}Upper} &
  HTC 4 &
  \multicolumn{1}{c|}{\cellcolor[HTML]{EFEFEF}{\color[HTML]{34A853} \textbf{74.7 (45)}}} &
  {\color[HTML]{34A853} \textbf{64.4 (106)}} &
  {\color[HTML]{EA4335} \textbf{64.8 (423)}} &
  \multicolumn{1}{c|}{\cellcolor[HTML]{EFEFEF}{\color[HTML]{34A853} \textbf{65.4 (322)}}} &
  {\color[HTML]{34A853} \textbf{68.8 (437)}} \\ \hline
\cellcolor[HTML]{EFEFEF} &
  \cellcolor[HTML]{EFEFEF}HTC 1 &
  \multicolumn{1}{c|}{{\color[HTML]{EA4335} \textbf{0 (0)}}} &
  63 (10) &
  {\color[HTML]{EA4335} \textbf{69 (52)}} &
  \multicolumn{1}{c|}{{\color[HTML]{34A853} \textbf{67.8 (61)}}} &
  72.2 (73) \\ \cline{2-2}
\rowcolor[HTML]{EFEFEF} 
\cellcolor[HTML]{EFEFEF} &
  HTC 2 &
  \multicolumn{1}{c|}{\cellcolor[HTML]{EFEFEF}41.2 (2)} &
  {\color[HTML]{EA4335} \textbf{62.8 (12)}} &
  {\color[HTML]{34A853} \textbf{73.2 (74)}} &
  \multicolumn{1}{c|}{\cellcolor[HTML]{EFEFEF}66.4 (83)} &
  72.4 (110) \\ \cline{2-2}
\cellcolor[HTML]{EFEFEF} &
  \cellcolor[HTML]{EFEFEF}HTC 3 &
  \multicolumn{1}{c|}{53.7 (8)} &
  66.4 (26) &
  73 (149) &
  \multicolumn{1}{c|}{{\color[HTML]{EA4335} \textbf{64.7 (161)}}} &
  {\color[HTML]{EA4335} \textbf{71 (212)}} \\ \cline{2-2}
\rowcolor[HTML]{EFEFEF} 
\multirow{-4}{*}{\cellcolor[HTML]{EFEFEF}Lower} &
  HTC 4 &
  \multicolumn{1}{c|}{\cellcolor[HTML]{EFEFEF}{\color[HTML]{34A853} \textbf{66.8 (14)}}} &
  {\color[HTML]{34A853} \textbf{72.6 (54)}} &
  72.8 (304) &
  \multicolumn{1}{c|}{\cellcolor[HTML]{EFEFEF}67 (321)} &
  {\color[HTML]{34A853} \textbf{73 (440)}} \\ \hline
\end{tabular}%
}
\end{table}

\subsection{Numbering analysis on test data}
\label{section:numbering-analysis}

The numbering task consists in detecting all tooth instances and correctly classifying them.
This task has a direct practical value, as the radiologists must inform the patients' missing teeth in the reports.
Additionally to this practical application, numbering is helpful to assess the HITL benefits on a task less sensitive to coarse predictions.

We evaluated the model performances according to their errors, which in the numbering task may be grouped in three types:
\begin{itemize}
    \item False negatives (when the model does not detect an instance);
    \item False positives (when the model detects something that is not an instance of an object of interest);
    \item Misclassifications (when the model correctly detects an object instance but classifies it wrongly).
\end{itemize}

We synthesize these values along with the total errors and the true positives for the trained networks in Table \ref{tab:numbering-performance} using a 0.5 IoU detection threshold.
One can perceive a consistent performance improvement trend in all values over the iterations.
The most significant advancement was over the false negatives, reduced by 63\%.
The misclassification errors shrank 27\%.

\begin{table}[t]
\centering
\caption{Neural network performances according to the error types on the numbering task for a 0.5 IoU detection threshold over the test data set. All errors shrank at each HITL iteration.}
\resizebox{\textwidth}{!}{%
\label{tab:numbering-performance}
\begin{tabular}{@{}cccccc@{}}
\toprule
Network & False Negatives & False Positives & Misclassifications & Total Errors & True Positives \\ \midrule
HTC 1 & 111 & 74 & 216 & 401 & 11,390 \\
HTC 2 & 85  & 53 & 179 & 317 & 11,453 \\
HTC 3 & 78  & 52 & 166 & 296 & 11,473 \\
HTC 4 & 41  & 42 & 158 & 241 & 11,518 \\ \bottomrule
\end{tabular}%
}
\end{table}

\begin{figure}
\begin{subfigure}{\textwidth}
  \centering
  \includegraphics[width=.8\textwidth]{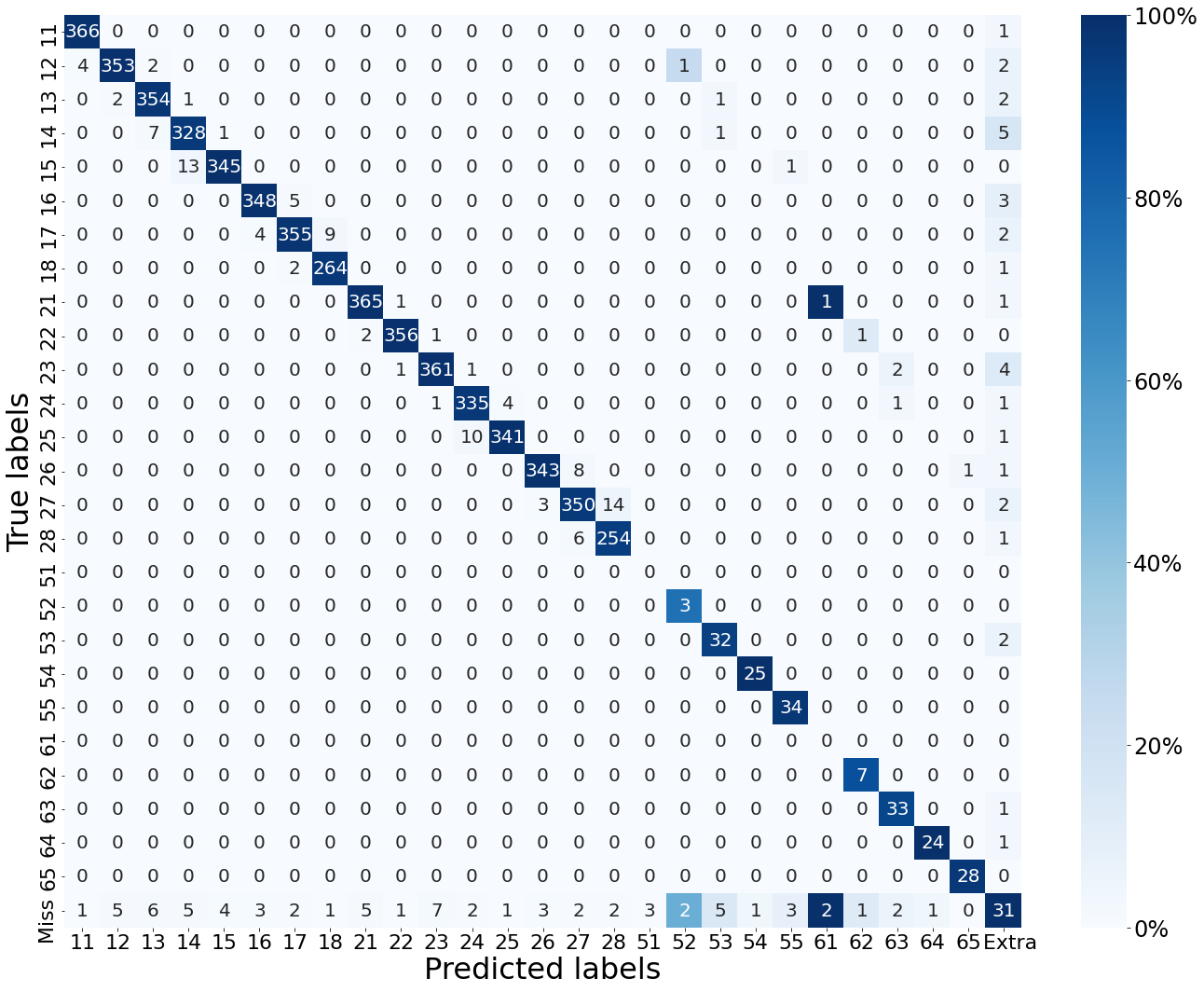}
  \caption{HTC 1's upper teeth confusion matrix.}
  \label{subfig:htc1-upper}
\end{subfigure}
\newline
\begin{subfigure}{\textwidth}
  \centering
  \includegraphics[width=.8\textwidth]{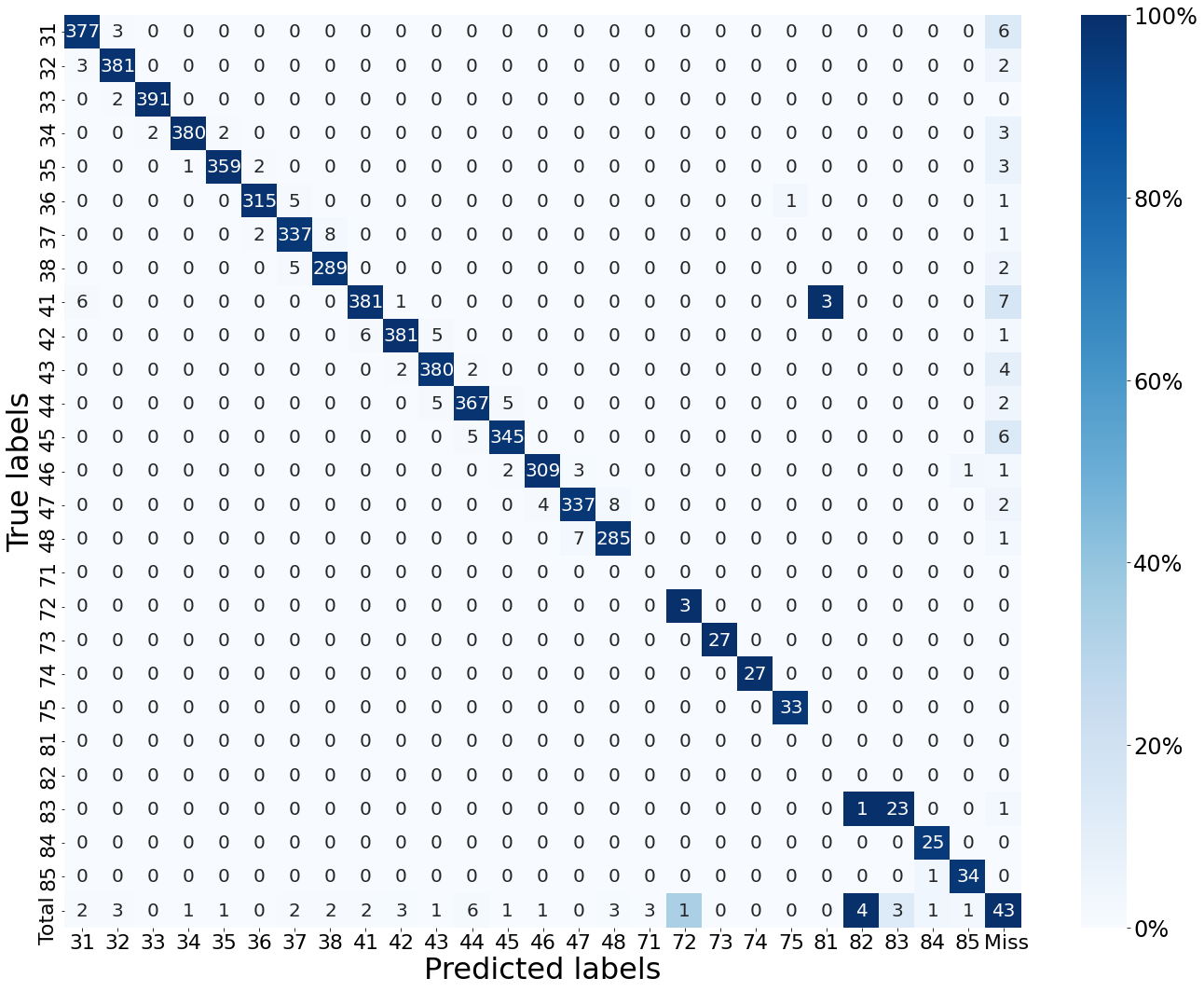}
  \caption{HTC 1's lower teeth confusion matrix.}
  \label{subfig:htc1-lower}
\end{subfigure}
\caption{HTC 1's upper and lower teeth confusion matrices for a 0.5 IoU detection threshold. The last lines are for the false negatives per tooth type, while the last columns are for the false positives.}
\label{fig:confusion-matrix-htc1}
\end{figure}

\begin{figure}
\begin{subfigure}{\textwidth}
  \centering
  \includegraphics[width=.8\textwidth]{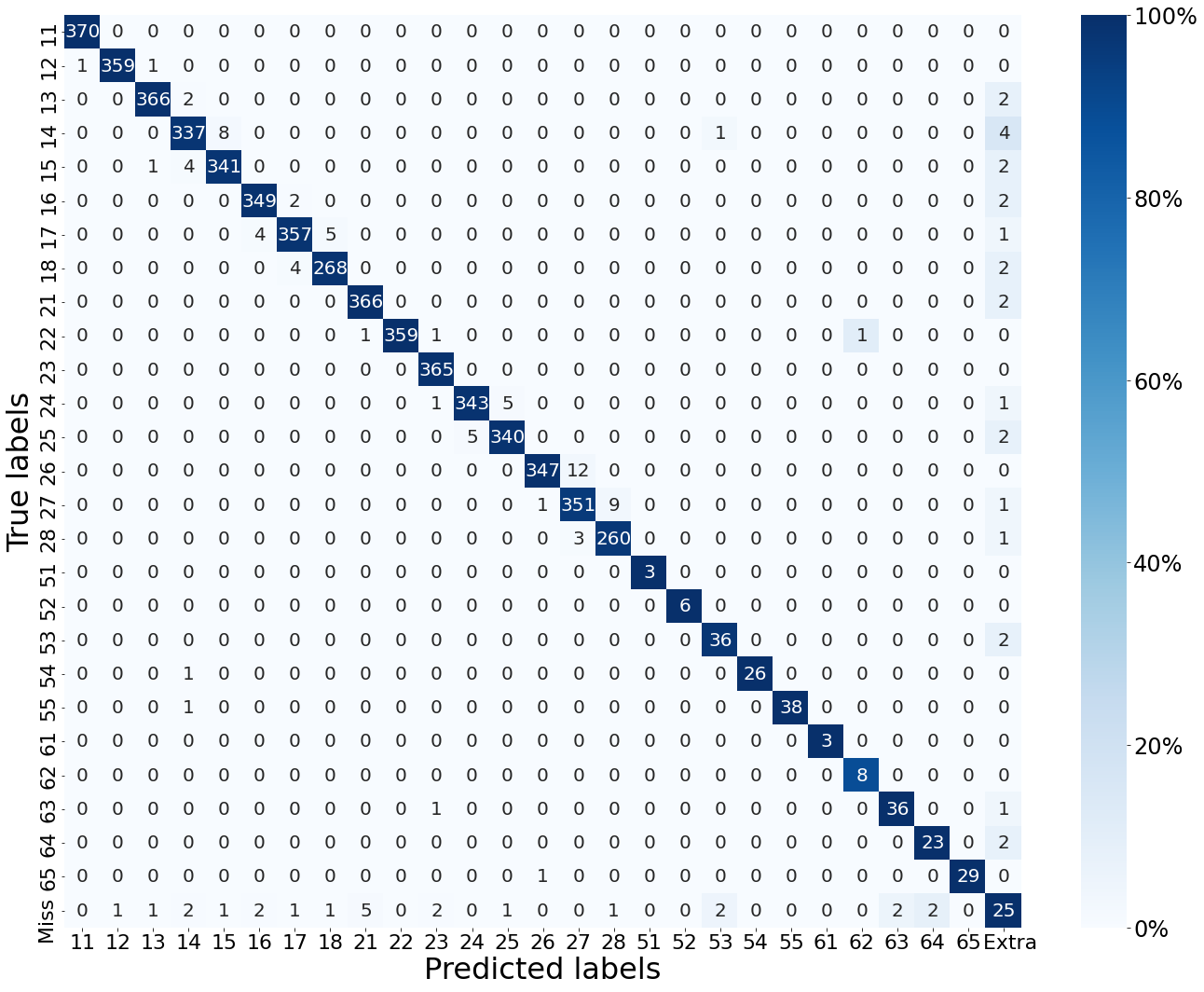}
  \caption{HTC 4's upper teeth confusion matrix.}
  \label{subfig:htc4-upper}
\end{subfigure}
\newline
\begin{subfigure}{\textwidth}
  \centering
  \includegraphics[width=.8\textwidth]{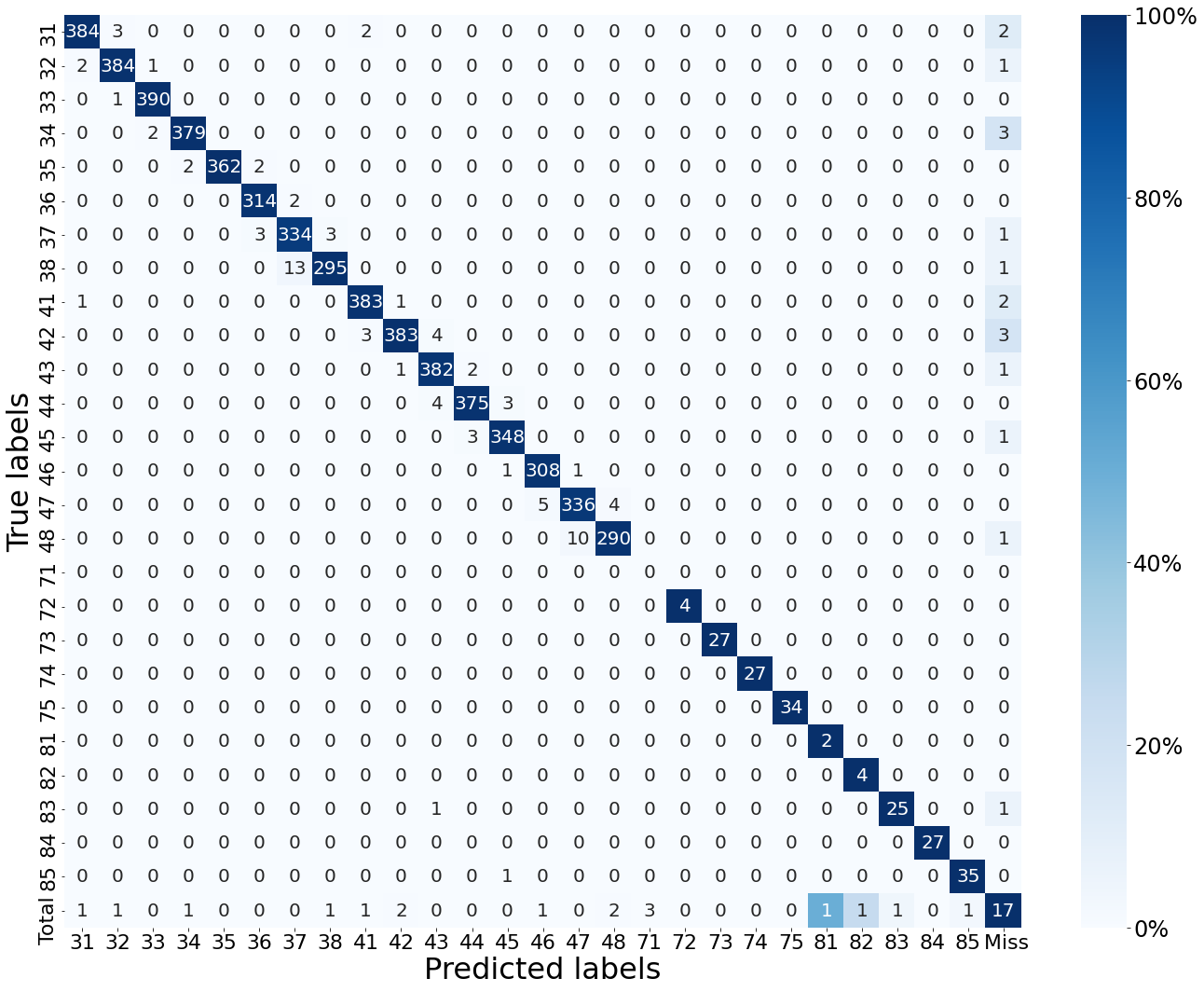}
  \caption{HTC 4's lower teeth confusion matrix.}
  \label{subfig:htc4-lower}
\end{subfigure}
\caption{HTC 4's upper and lower teeth confusion matrices for a 0.5 IoU detection threshold. The last lines are for the false negatives per tooth type, while the last columns are for the false positives.}
\label{fig:confusion-matrix-htc4}
\end{figure}

The aggregated results from Table \ref{tab:numbering-performance} do not allow a detailed analysis of the numbering errors.
To solve that, we plotted the confusion matrices according to tooth types.
For brevity, we depict only HTC 1's and HTC 4's detection confusion matrices in Figures \ref{fig:confusion-matrix-htc1} and \ref{fig:confusion-matrix-htc4}, in which we split the matrices into the upper teeth and lower teeth parts for visualization purposes.
This division is not harmful to the analyses because misclassifications between those groups are rare.
A performance boost can be observed in all tooth groups by comparing HTC 1's and HTC4's confusion matrices.
The upper teeth were slightly easier to detect and classify for both networks.
The deciduous teeth were more challenging to detect correctly but easier to classify than permanent ones.
The misclassifications were essentially among nearby, same-function teeth, especially the premolars and molars.
The numbering of premolars and molars may be quite challenging in some circumstances, such as on unhealthy missing-tooth mouths, where dubious situations may occur even for human experts.

\subsection{Labeling time analysis}
\label{section:labeling-time-analysis}

\begin{figure}[t]
    \centering
    \includegraphics[width=\textwidth]{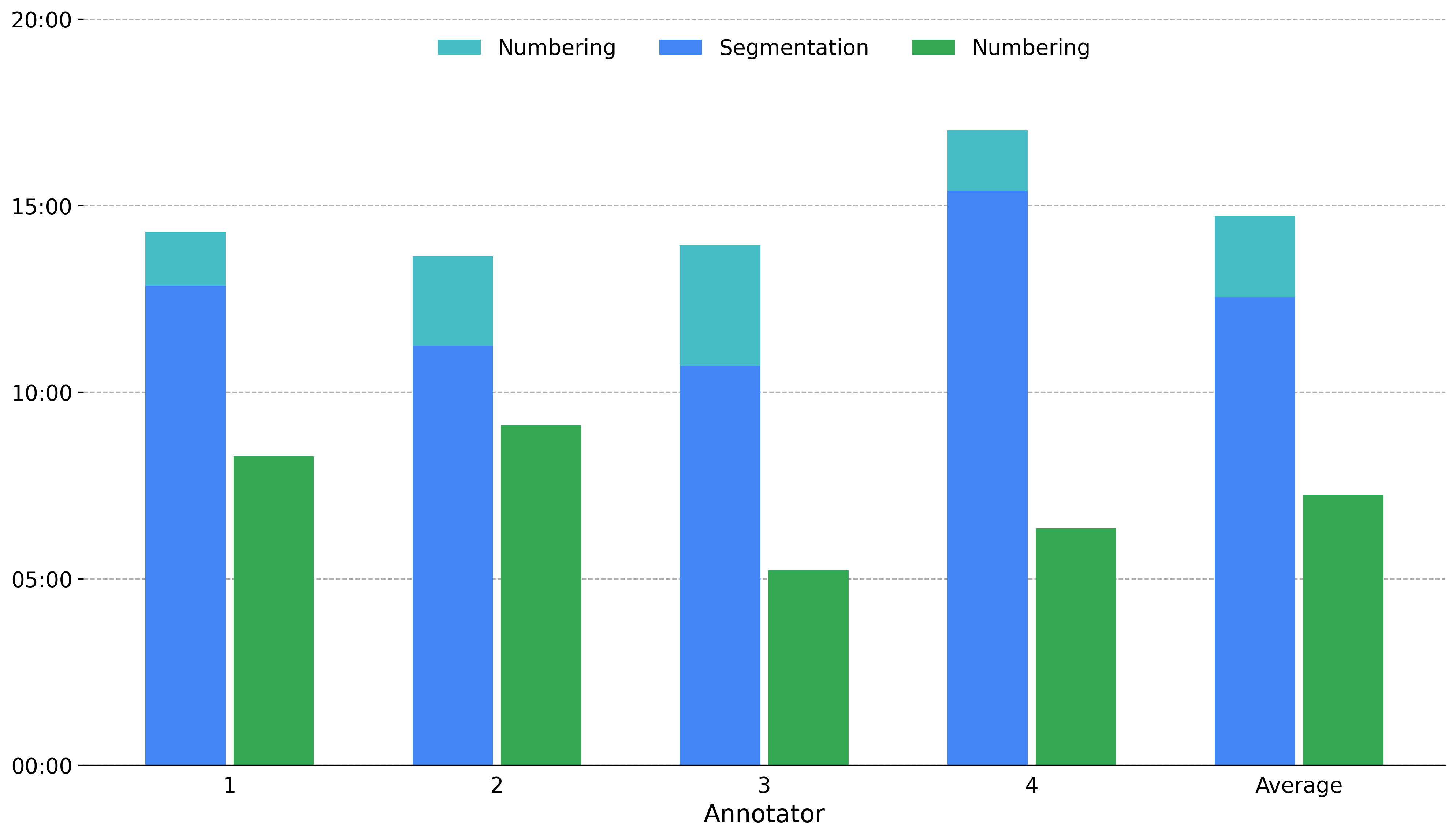}
    \caption{Comparison of each annotator time for HITL labeling verification time against manual labeling time, the latter split into segmentation and tooth numbering. The HITL labeling lasts considerably less (51\%) than manual labeling.}
    \label{fig:hitl-times}
\end{figure}

In a HITL setup, we are not only interested in labeling quality, but also labeling speed-up.
Therefore, we monitored the HITL labeling verification and the radiograph manual labeling times.
Figure \ref{fig:hitl-times} compares those two labeling approaches according to each annotators' average time.
These time values were measured during the third iteration, in which we asked our annotators to clock their correction and manual labeling.
For the manual labeling, they split their time into segmentation and tooth numbering.
The latter is the time to type and assign the tooth class.

From Figure \ref{fig:hitl-times}, one can perceive that the labeling verification procedure was significantly faster than manual labeling.
Labeling radiographs manually lasted on average 14 minutes and 43 seconds per radiograph, while labeling using the HITL concept took 14 minutes and 43 seconds, a 51\% time reduction.
The annotation verification was faster than manual segmentation, even if we disregarded the numbering procedure.
In that case, the HITL approach reduced the labeling time by 42\% compared to manual labeling, which took 12 minutes and 33 seconds on average.
If we considered the 51\% time reduction, the HITL procedure saved more than 390 continuous working hours.

\subsection{HITL bottlenecks}
\label{section:bottlenecks}

We investigated possible bottlenecks that could have significantly slowed down the HITL verification procedure.
Already in the first iteration, in which we established the verification protocol, our annotators mentioned several times the presence of serrated segmentation that comprises most of the correction time.
This serrated pattern came from the $28 \times 28$ low-resolution masks and appeared especially on the tooth crowns, but it was also frequent on the other parts of large and complex-shaped tooth instances, such as molars.
The annotators with no deep learning background considered these incongruous masks somewhat surprising, as the crowns are usually well-defined and easier for humans to segment.
For segmentation models, the tooth crowns are fine-detailed objects with acute borders, which hampers the segmentation task.

In order to understand how much impact these jagged contours have in the HITL, we quantified the correction fractions related to tooth parts: Crown, middle, root, or a combination of them, including correction on all tooth parts.
We automatically split each tooth instance into these three parts in the vertical axis for this analysis and measure the frequency of the modifications in each part, disregarding the size of the changes.

\begin{figure}[t]
    \centering
    \includegraphics[width=\textwidth]{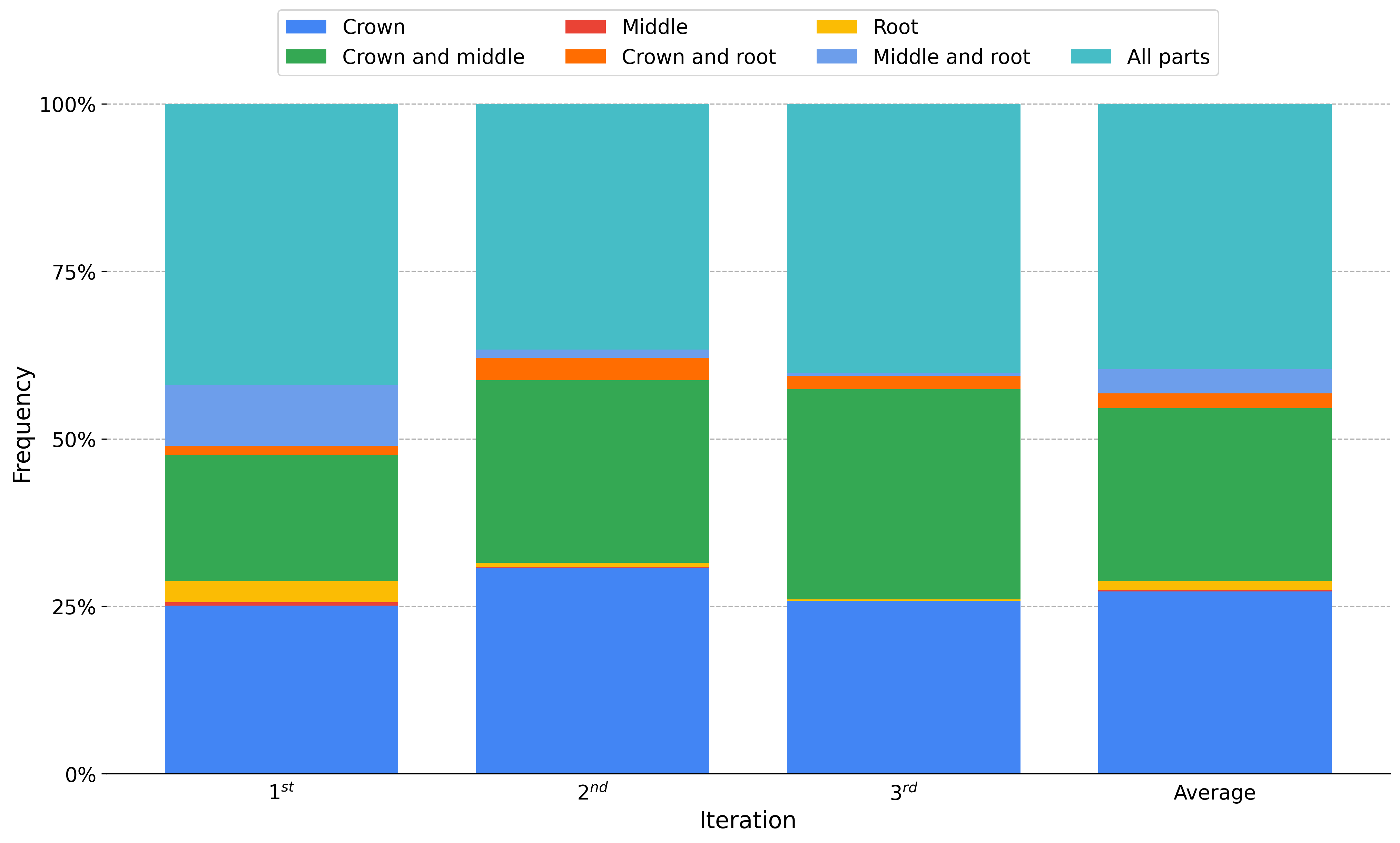}
    \caption{Frequency of corrections according to tooth part. The crown segmentation was adjusted in more than 85\% of the corrected instances in all HITL iterations.}
    \label{fig:frequency-corrections}
\end{figure}

Figure \ref{fig:frequency-corrections} summarizes the obtained results, showing the frequency of parts where the correction took place at each iteration.
One can perceive that, in all iterations, adjustments in the crown segmentation have been made in more the 85\% of the corrected instances.
These adjustments were highly frequent, mainly due to the serrated patterns and heavily slowed down the verification process.
Possible solutions to reduce this issue when neglecting these tiny errors is not an option include increasing the segmentation mask resolution, employing a two-stage instance segmentation approach, or using a more specialized method, such as the PointRend module.

\subsection{Qualitative analysis}
\label{section:qualitative-analysis}

The quantitative analyses guided our qualitative analyses.
We focused on the best and worst results according to the primary metric, comparing the ground truth with the network predictions and the verified labels.
Figures \ref{fig:qualitative-results} (a) and (b) illustrate the best and the worst HTC 4's results, respectively, according to the segmentation mAP on the test data set.
Figure \ref{fig:qualitative-results} (a) corresponds to a well-focused, crisp and clear radiograph from a 32-teeth healthy mouth, characteristics common to the best results.
From the zoomed area, we see that the annotation correction led to a final label closer to the ground truth and also less noisy.

The worst result, illustrated in Figure \ref{fig:qualitative-results} (b), came from a slightly blurry image from an unhealthy mouth, a common pattern in the radiographs of the worst results.
However, in this case, the network performance was reasonably good, and the low metric was due to main factors.
First and most important, the annotator wrongly labeled the teeth 32, 33, 34 and 35 respectively as teeth 31, 32, 33 and 34, probably due to a sequence of typos, which reduced the segmentation mAP significantly.
Second, the presence of radiolucent material prostheses and restoration encumbered the segmentation task for both model and annotator.
The zoomed area shows that model undersegmented those spots, which were adjusted by the annotator, but still missed some areas.
The other annotation corrections smoothed the noisy borders and reduced the difference from the ground truth labels.
We additionally illustrate in Figure \ref{fig:qualitative-results} (c) a sample result on a mixed dentition mouth.
These radiographs are challenging for models and human annotators due to overlapping.
In this particular image, there are also occlusions between posterior teeth, hardening the task.
However, the model prediction proved to be adequate, even before the labeling verification.
The zoomed area shows that the corrections reduced the gap to the manual ground truth labels, but there were some divergences for root segmentation of teeth 54 and 55.

\begin{figure}
\begin{subfigure}{\textwidth}
    \centering
    \includegraphics[width=\textwidth]{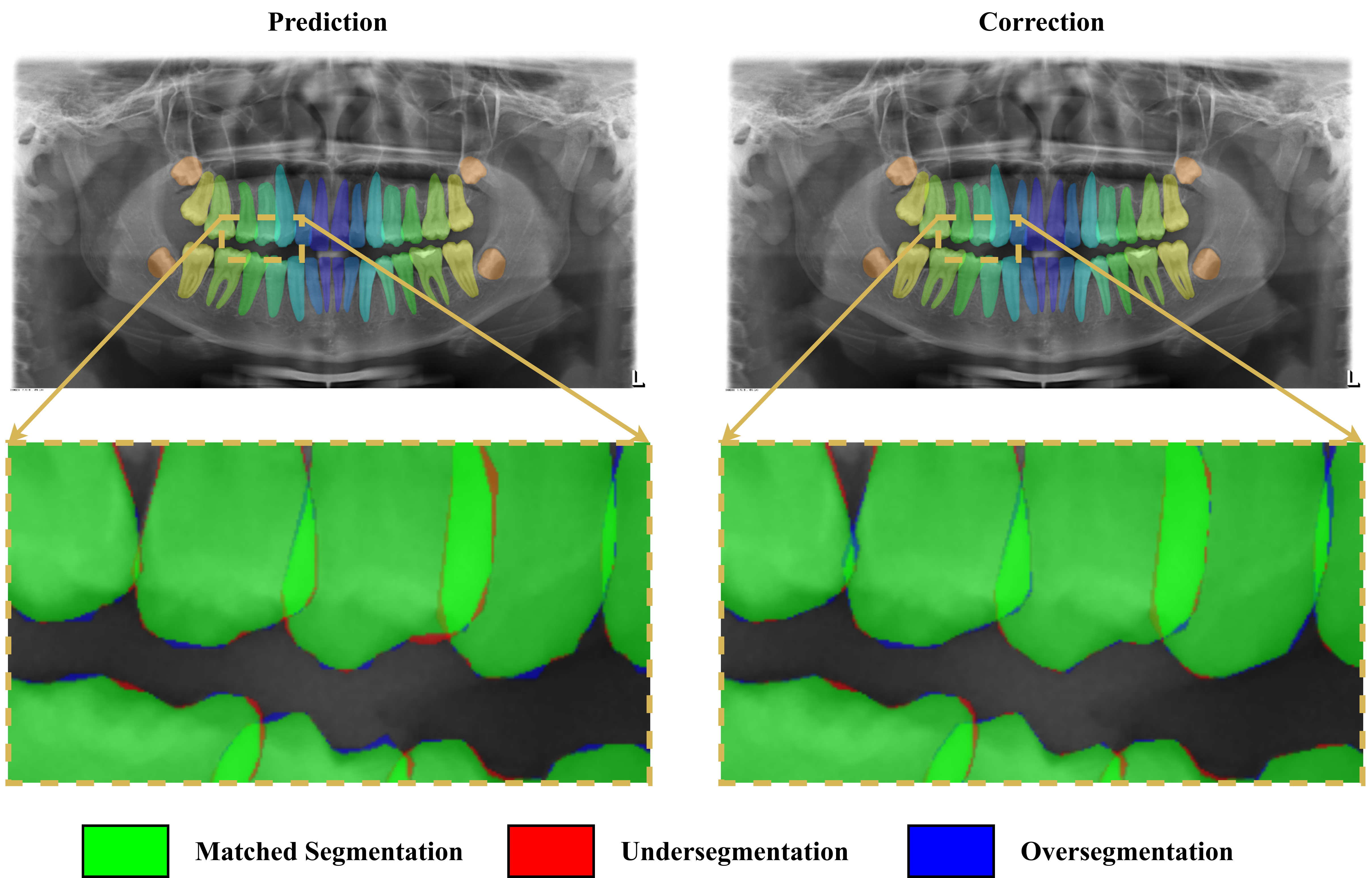}
    \caption{HTC 4's best result, which happened on a well-focused, crisp and clear radiograph from a 32-teeth healthy mouth, characteristics common to the best results.}
    \label{subfig:htc4-best}
\end{subfigure}
\newline
\begin{subfigure}{\textwidth}
    \centering
    \includegraphics[width=\textwidth]{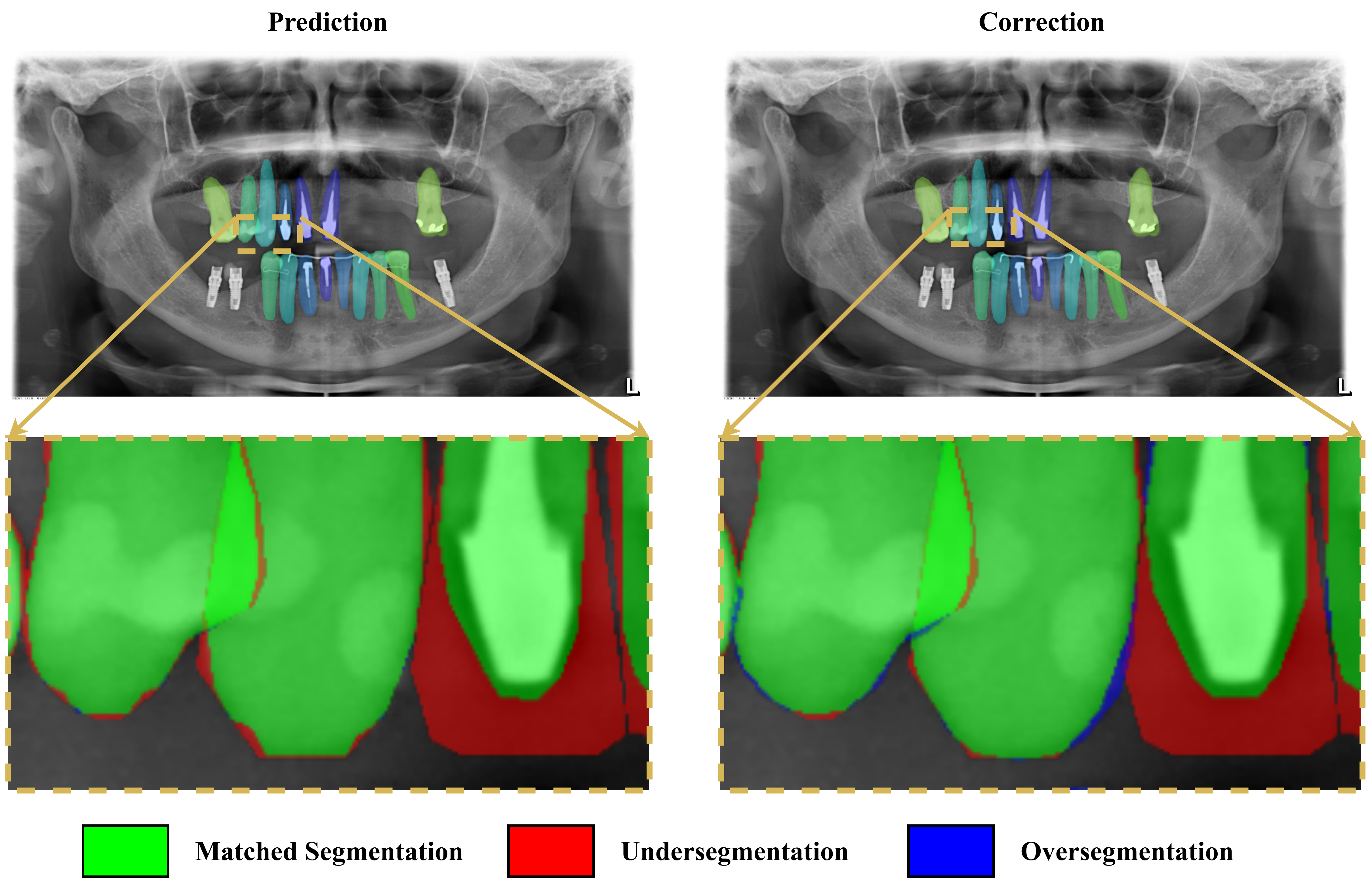}
    \caption{HTC 4's worst result, which happened on a slightly blurry radiograph from an unhealthy mouth with radiolucent material prostheses.}
    \label{subfig:htc4-worst}
\end{subfigure}
\end{figure}

\begin{figure}[t]\ContinuedFloat
\begin{subfigure}{\textwidth}
    \centering
    \includegraphics[width=\textwidth]{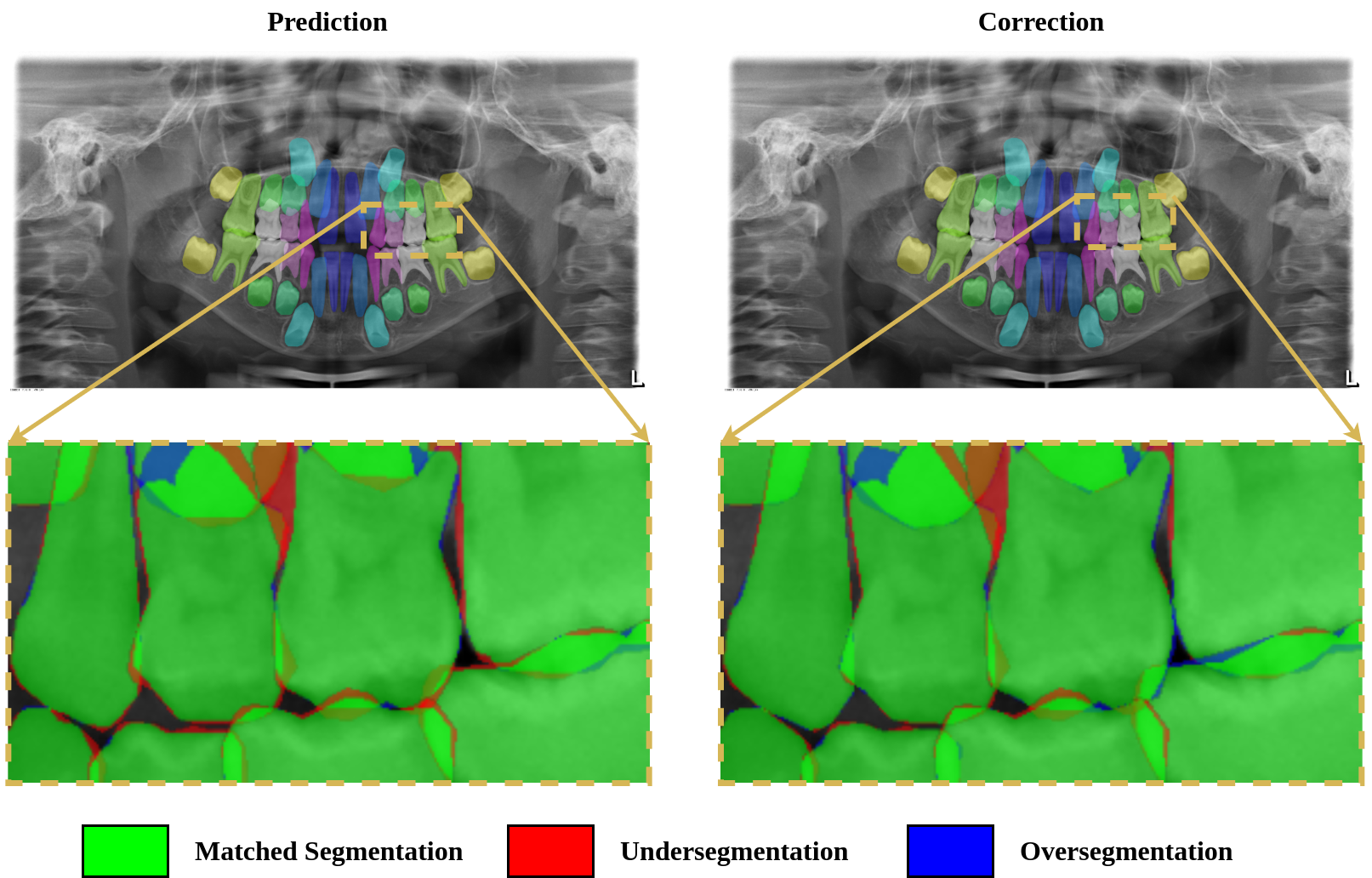}
    \caption{Sample of HTC 4's results on a mixed dentition radiograph.}
    \label{subfig:htc4-mixed}
\end{subfigure}
\caption{HTC 4's best and worst results according to the segmentation mAP on the test set, and an additional result sample on a mixed dentition radiograph. The illustrations compare the predictions before and after the corrections by the annotators. The zoomed areas highlight the matched segmentation, undersegmentation, and oversegmentation with the ground truth labels, evincing that the corrections led the final labels to be less noisy and closer to the ground truth.}
\label{fig:qualitative-results}
\end{figure}

\section{Submission platform, evaluation protocols, and baselines}

Our data set comprises 4000 labeled radiographs (850 manually labeled and 3150 HITL labeled), from which 2000 are used for solution assessments in the \textbf{OdontoAI platform}\footnote{The platform link will be available upon the article's acceptance and publication.}.
The platform consists of a website where researchers can submit their predictions in a standardized fashion, enabling a fair benchmarking of the proposed methods.
We provide 2000 radiographs along with their labels (650 manually labeled and 1350 HITL labeled radiographs) for model training and validation.
The remaining 2000 images (1800 labeled in the HITL scheme and 200 manually labeled) do not have their labels publicly available and consist of the platform test set.
We also provide in the platform precise instructions on how to submit solutions and open-source codes of the used metrics and for creating the submission files.

We configured three benchmarks for the OdontoAI platform, comprising classical computer vision tasks useful for analyzing dental panoramic radiographs.
The tasks are tooth \textbf{instance segmentation}, \textbf{semantic segmentation}, and \textbf{numbering}, which we detail in the following sections together with the selected metrics.
As baselines, we included the results of neural networks trained on the 2000 publicly available labeled radiographs architectures using the architectures presented in our instance segmentation benchmark (Section \ref{section:benchmark}).

\subsection{Instance segmentation task}
\label{sec:inst-segm-task}

The instance segmentation task is a straightforward application of our data set.
This task is challenging and comprehensive, as it combines instance detection and segmentation.
Many researchers investigate instance segmentation due to its usefulness, but lack of data may be an issue.
Our data set solves this problem.

We chose mAP as the main metric to evaluate instance segmentation.
A rigid metric is necessary, as our experiments showed that the AP50 and AP75 are rather loose metrics to the task.
The adopted metric, mAP, is not only stricter but also more comprehensive, being suitable for the instance segmentation task benchmark of the OdontoAI platform.
AP50 and AP75 are included as secondary metrics as well the equivalent metrics for detection with bounding boxes.
Table \ref{tab:inst-segm-baselines} illustrates a sample of the benchmark ranking available in our platform, with the attained results by the baselines.

\begin{table}[t]
\centering
\caption{A sample of the OdontoAI platform benchmark ranking for the instance segmentation task with baselines.}
\label{tab:inst-segm-baselines}
\resizebox{\textwidth}{!}{%
\begin{tabular}{|c|l|ccc|ccc|}
\hline
\rowcolor[HTML]{EFEFEF} 
\cellcolor[HTML]{EFEFEF} &
  \cellcolor[HTML]{EFEFEF} &
  \multicolumn{3}{c|}{\cellcolor[HTML]{EFEFEF}Detection} &
  \multicolumn{3}{c|}{\cellcolor[HTML]{EFEFEF}Segmentation} \\ \cline{3-8} 
\rowcolor[HTML]{EFEFEF} 
\multirow{-2}{*}{\cellcolor[HTML]{EFEFEF}Rank} &
  \multirow{-2}{*}{\cellcolor[HTML]{EFEFEF}Architecture} &
  \multicolumn{1}{c|}{\cellcolor[HTML]{EFEFEF}AP50} &
  \multicolumn{1}{c|}{\cellcolor[HTML]{EFEFEF}AP75} &
  mAP &
  \multicolumn{1}{c|}{\cellcolor[HTML]{EFEFEF}AP50} &
  \multicolumn{1}{c|}{\cellcolor[HTML]{EFEFEF}AP75} &
  mAP \\ \hline
1 &
  HTC &
  \multicolumn{1}{c|}{{\color[HTML]{34A853} \textbf{0.924}}} &
  \multicolumn{1}{c|}{0.964} &
  {\color[HTML]{34A853} \textbf{0.821}} &
  \multicolumn{1}{c|}{{\color[HTML]{34A853} \textbf{0.941}}} &
  \multicolumn{1}{c|}{0.964} &
  {\color[HTML]{34A853} \textbf{0.821}} \\ \cline{1-2}
\rowcolor[HTML]{EFEFEF} 
2 &
  DetectoRS &
  \multicolumn{1}{c|}{\cellcolor[HTML]{EFEFEF}0.920} &
  \multicolumn{1}{c|}{\cellcolor[HTML]{EFEFEF}{\color[HTML]{34A853} \textbf{0.967}}} &
  0.803 &
  \multicolumn{1}{c|}{\cellcolor[HTML]{EFEFEF}0.933} &
  \multicolumn{1}{c|}{\cellcolor[HTML]{EFEFEF}{\color[HTML]{34A853} \textbf{0.967}}} &
  0.809 \\ \cline{1-2}
3 &
  Cascade R-CNN &
  \multicolumn{1}{c|}{0.893} &
  \multicolumn{1}{c|}{0.951} &
  0.786 &
  \multicolumn{1}{c|}{0.920} &
  \multicolumn{1}{c|}{0.952} &
  0.790 \\ \cline{1-2}
\rowcolor[HTML]{EFEFEF} 
4 &
  Cascade R-CNN with DCN &
  \multicolumn{1}{c|}{\cellcolor[HTML]{EFEFEF}0.875} &
  \multicolumn{1}{c|}{\cellcolor[HTML]{EFEFEF}0.930} &
  0.773 &
  \multicolumn{1}{c|}{\cellcolor[HTML]{EFEFEF}0.886} &
  \multicolumn{1}{c|}{\cellcolor[HTML]{EFEFEF}0.931} &
  0.770 \\ \cline{1-2}
5 &
  Mask R-CNN &
  \multicolumn{1}{c|}{0.868} &
  \multicolumn{1}{c|}{0.935} &
  0.749 &
  \multicolumn{1}{c|}{0.893} &
  \multicolumn{1}{c|}{0.936} &
  0.760 \\ \cline{1-2}
\rowcolor[HTML]{EFEFEF} 
6 &
  ResNeSt Cascade R-CNN &
  \multicolumn{1}{c|}{\cellcolor[HTML]{EFEFEF}0.866} &
  \multicolumn{1}{c|}{\cellcolor[HTML]{EFEFEF}0.903} &
  0.764 &
  \multicolumn{1}{c|}{\cellcolor[HTML]{EFEFEF}0.849} &
  \multicolumn{1}{c|}{\cellcolor[HTML]{EFEFEF}0.903} &
  0.652 \\ \cline{1-2}
7 &
  ResNeSt Mask R-CNN &
  \multicolumn{1}{c|}{{\color[HTML]{EA4335} \textbf{0.825}}} &
  \multicolumn{1}{c|}{{\color[HTML]{EA4335} \textbf{0.879}}} &
  {\color[HTML]{EA4335} \textbf{0.726}} &
  \multicolumn{1}{c|}{{\color[HTML]{EA4335} \textbf{0.828}}} &
  \multicolumn{1}{c|}{{\color[HTML]{EA4335} \textbf{0.880}}} &
  {\color[HTML]{EA4335} \textbf{0.637}} \\ \hline
\end{tabular}%
}
\end{table}

\subsection{Semantic segmentation task}
\label{sec:seman-segm-task}

Semantic segmentation is also a basilar task in computer vision, being the reason why we included it in our platform's benchmarks.
The tasks consist of segmenting classes precisely as possible, disregarding object instances.
In our benchmark, there is only one class (tooth), and the researchers should propose methods to distinguish it from the background.
Due to this dichotomic nature, we employed the usual metrics for binary segmentation: accuracy, specificity, precision, recall, f1-score, and IoU.
The latter is the main metric, and it is equivalent to the binary mIoU, a commonly used metric in semantic segmentation benchmarks.

Table \ref{tab:semantic-segm-baselines} shows a sample of the benchmark ranking at the OdontoAI platform for the semantic segmentation task.
The baseline metrics were computed after converting the instance segmentation predictions of the networks into segmentation masks.

\begin{table}[t]
\centering
\caption{A sample of the OdontoAI platform benchmark ranking for the semantic segmentation task with baselines.}
\label{tab:semantic-segm-baselines}
\resizebox{\textwidth}{!}{%
\begin{tabular}{|c|l|c|c|c|c|c|c|}
\hline
\rowcolor[HTML]{EFEFEF} 
Rank &
  \multicolumn{1}{c|}{\cellcolor[HTML]{EFEFEF}Architecture} &
  Accuracy (\%) &
  Specificity (\%) &
  Precision (\%) &
  Recall (\%) &
  F1-score (\%) &
  IoU (\%) \\ \hline
1 &
  HTC &
  {\color[HTML]{34A853} \textbf{98.8}} &
  {\color[HTML]{34A853} \textbf{99.5}} &
  {\color[HTML]{34A853} \textbf{98.2}} &
  {\color[HTML]{34A853} \textbf{96.2}} &
  {\color[HTML]{34A853} \textbf{97.2}} &
  {\color[HTML]{34A853} \textbf{94.5}} \\ \hline
\rowcolor[HTML]{EFEFEF} 
2 & DetectoRS              & 98.7                                 & 99.4                                 & 97.8 & 96.1                                 & 96.9 & 94.1 \\ \hline
3 & Cascade R-CNN          & 98.7                                 & 99.4                                 & 97.7 & 96.1                                 & 96.9 & 94.0 \\ \hline
\rowcolor[HTML]{EFEFEF} 
4 & Cascade R-CNN with DCN & 98.7                                 & {\color[HTML]{34A853} \textbf{99.5}} & 98.0 & 95.6                                 & 96.8 & 93.8 \\ \hline
5 & Mask R-CNN             & 98.6                                 & 99.3                                 & 97.4 & 95.9                                 & 96.6 & 93.5 \\ \hline
\rowcolor[HTML]{EFEFEF} 
6 & ResNeSt Cascade R-CNN  & {\color[HTML]{EA4335} \textbf{97.0}} & 98.6                                 & 94.7 & {\color[HTML]{EA4335} \textbf{91.2}} & 92.9 & 86.7 \\ \hline
7 &
  ResNeSt Mask R-CNN &
  {\color[HTML]{EA4335} \textbf{97.0}} &
  {\color[HTML]{EA4335} \textbf{98.5}} &
  {\color[HTML]{EA4335} \textbf{94.3}} &
  91.3 &
  {\color[HTML]{EA4335} \textbf{92.8}} &
  {\color[HTML]{EA4335} \textbf{86.5}} \\ \hline
\end{tabular}%
}
\end{table}

\subsection{Numbering task}
\label{sec:numbering-task}

Finally, we included the task of ``numbering," which is almost equivalent to the multi-label classification computer vision task.
It slightly differs from the multi-label classification task because one or more supernumerary teeth may appear.
In the numbering task of our benchmark, the goal is to predict the present teeth in the panoramic radiograph.
Although this task may not be advantageous as a preprocessing step for analyzing panoramic radiographs, it naturally appears in practical applications such as form fillings and automatic report generation.
While reports customary document the patient's permanent missing teeth, the OdontoAI platform's numbering task expects a list of present teeth.
We chose this conventional because deciduous and supernumerary teeth may occur.

Our experiments showed that it is easy to identify the present teeth correctly.
Through a general instance segmentation (HTC 4 neural network), the numbering task resulted in only 241 errors among false positives, false negatives, and misclassifications.
Therefore, we chose a rather rigorous metric main metric, ``exact match,'' in which a true positive is only taken into account when all tooth numbering predictions are correct.
Other than the main metric, the OdontoAI platform includes other metrics, such as micro accuracy, micro precision, micro recall and Hamming loss.
The Hamming loss averages the fraction of the incorrect predictions for each label.
We illustrate in Table \ref{tab:numbering-baselines} a sample of numbering benchmark ranking found at the OdontoAI platform.

\begin{table}[t]
\centering
\caption{A sample of the OdontoAI platform benchmark ranking for the numbering task with baselines.}
\label{tab:numbering-baselines}
\resizebox{\textwidth}{!}{%
\begin{tabular}{|c|l|c|c|c|c|c|}
\hline
\rowcolor[HTML]{EFEFEF} 
Rank &
  \multicolumn{1}{c|}{\cellcolor[HTML]{EFEFEF}Architecture} &
  Exact Match (\%) &
  Micro Accuracy (\%) &
  Micro Precision (\%) &
  Micro Recall (\%) &
  Hamming Loss \\ \hline
1 &
  HTC &
  {\color[HTML]{34A853} \textbf{67.9}} &
  {\color[HTML]{34A853} \textbf{98.6}} &
  {\color[HTML]{34A853} \textbf{98.8}} &
  {\color[HTML]{34A853} \textbf{98.5}} &
  {\color[HTML]{34A853} \textbf{0.0143}} \\ \hline
\rowcolor[HTML]{EFEFEF} 
2 & DetectoRS              & 66.2 & 98.4 & 98.7 & 98.3 & 0.0164 \\ \hline
3 & Cascade R-CNN with DCN & 65.7 & 98.4 & 98.7 & 98.3 & 0.0161 \\ \hline
\rowcolor[HTML]{EFEFEF} 
4 & Cascade R-CNN          & 62.8 & 98.2 & 98.5 & 98.2 & 0.0177 \\ \hline
5 & ResNeSt Cascade R-CNN  & 60.7 & 97.9 & 98.4 & 97.8 & 0.0206 \\ \hline
\rowcolor[HTML]{EFEFEF} 
6 & Mask R-CNN             & 59.0 & 98.0 & 98.5 & 97.8 & 0.0197 \\ \hline
7 &
  ResNeSt Mask R-CNN &
  {\color[HTML]{EA4335} \textbf{56.3}} &
  {\color[HTML]{EA4335} \textbf{97.7}} &
  {\color[HTML]{EA4335} \textbf{98.4}} &
  {\color[HTML]{EA4335} \textbf{97.3}} &
  {\color[HTML]{EA4335} \textbf{0.0231}} \\ \hline
\end{tabular}%
}
\end{table}

\section{Discussion and concluding remarks}

In this work, we constructed a large-size labeled data set of dental panoramic radiographs: \textbf{the OdontoAI Open Panoramic Radiographs (O$^2$PR) data set}.
The O$^2$PR comprises 4000 images, in which the teeth were segmented and numbered, and is four times larger than the previously most extensive data set on the matter available in the literature (\citeauthor{panetta2021tufts}, \citeyear{panetta2021tufts}).
The labels of 2,000 radiographs of the O$^2$PR data set are publicly available, while the labels of the remaining others remain private for solution assessments at our online platform, the \textbf{OdontoAI platform}.
We hope our platform along with our data set will boost computer vision research on dental panoramic radiographs as it enables a fair comparison of the proposed methods while serving as a task central to researchers in the field.

The magnitude of the (O$^2$PR) data set was attained through the use of the HTIL concept to speed-up the labeling process.
Our results indicated about 51\% of labeling time reduction, even instructing our annotators to attend to tiny segmentation errors.
We estimate having saved at least 390 continuous working hours.
In practice, this number is even bigger, as manual labeling is more human demanding.
The HITL annotation verification process is less burdensome, as confirming the labels through visual inspection (rather than correcting them through mouse clicks and the point drag-and-drop feature) corresponds to a significant fraction of the verification process.

The performance of the trained networks on distinct data (validation data, HITL data, and, most important, on a separated manually labeled test data set) were consistent, showing an increasing trend in the considered metrics over the HITL iterations.
HTC 4's segmentation mAP was +5.4 percentage points higher than the HTC 1's on the test data set.
For comparison, the performance gain from the standard Mask R-CNN choice to the HTC, the winner architecture of our benchmark (Section \ref{section:benchmark}), was +4.4 in terms of segmentation mAP.
The work by \citeauthor{pinheiro2021numbering} (\citeyear{pinheiro2021numbering}) boosted the segmentation mAP of the Mask R-CNN architecture in +2 percentage points by replacing the original FCN segmentation head for a PointRend module.
This results reinforces a common, though frequently ignored, knowledge in the deep learning field: it is often better to gather data than expending much time in refining a model.
For the purpose of enlarging the labeled data sets, the HITL concept is very beneficial.

The less refined segmentation, mainly over the tooth crowns, was the major bottleneck for faster labeling.
The segmentation of deep learning solutions slightly differs from the human annotators, overall on the object's fine-detailed borders.
If the application allows neglecting these errors, the labeling speed up can increase substantially.
Therefore, we conclude that the HITL benefits for instance segmentation applications might vary significantly according to the application due to today's state of deep learning.
Currently, the HITL use is more beneficial to applications that do not demand greater segmentation accuracy than those that demand.
In this work, we did not neglect the tiny segmentation errors, as we want our data set to be general-purpose.

As future projects, we plan to extend the objects of interest (implants, prostheses, jaws) rather than increasing the data set size.
New publicly available data sets of dental panoramic X-rays from different devices, regardless of their sizes, will always be welcome and valuable resources for assessing the techniques' generalization capacity.
Future work on deep learning includes topics such as precise segmentation and tooth numbering considering the global context and geometrical relationships.

\bibliography{mybibfile}

\end{document}